%% file: main.tex
\documentclass{article} %
\usepackage{iclr2026_conference,times}

\input{math_commands.tex}

\usepackage{hyperref}
\usepackage{url}
\usepackage{booktabs} 
\usepackage{graphicx}
\usepackage{subcaption}
\usepackage{longtable}
\usepackage{array}
\usepackage{listings}
\usepackage{tcolorbox} 
\usepackage{siunitx}
\newcommand{\VarSty}[1]{\textnormal{\ttfamily\color{blue!90!black}#1}\unskip}

\definecolor{mygreen}{HTML}{3cb44b}
\definecolor{skyblue}{HTML}{beffff}
\definecolor{lightgreen}{HTML}{90ee90}
\definecolor{emerald}{rgb}{0.31, 0.78, 0.37}
\definecolor{mygreen}{HTML}{3cb44b}
\colorlet{myyellow}{green!10!orange!90!}
\definecolor{bgcolor}{rgb}{0.95, 0.95, 0.95}

\usepackage{amsmath,amssymb}

\usepackage{algorithm}
\usepackage{algpseudocode}  %

\usepackage{float}
\usepackage[table]{xcolor}

\newcommand{\pos}[1]{\cellcolor{green!18}{#1}}
\newcommand{\nega}[1]{\cellcolor{red!15}{#1}}

\title{LikeBench: Evaluating Subjective Likability in LLMs for Personalization}

\author{
Md Awsafur Rahman\thanks{Work done during an internship at Amazon.} \\
UC Santa Barbara \\
\texttt{awsaf@ucsb.edu}
\And
Adam Gabrys \\
Amazon \\
\texttt{gabrysa@amazon.com}
\And
Doug Kang \\
Amazon \\
\texttt{dougkang@amazon.com}
\And
Jingjing Sun \\
Amazon \\
\texttt{jingjins@amazon.com}
\And
Tian Tan \\
Amazon \\
\texttt{tianta@amazon.com}
\And
Ashwin Chandramouli \\
Amazon \\
\texttt{ashwic@amazon.com}
}

\iclrfinalcopy %
\begin{document}

\maketitle

\begin{abstract}
A personalized LLM should remember user facts, apply them correctly, and adapt over time to provide responses that the user prefers. Existing LLM personalization benchmarks are largely centered on two axes: accurately recalling user information and accurately applying remembered information in downstream tasks. We argue that a third axis, likability, is both subjective and central to user experience, yet under-measured by current benchmarks. To measure likability holistically, we introduce LikeBench\footnote{Code \& Data is planned for future release}, a multi-session, dynamic evaluation framework that measures likability across multiple dimensions by how much an LLM can adapt over time to a user’s preferences to provide more likable responses. In LikeBench, the LLMs engage in conversation with a simulated user and learn preferences only from the ongoing dialogue. As the interaction unfolds, models try to adapt to  responses, and after each turn, they are evaluated for likability across seven dimensions by the same simulated user. To the best of our knowledge, we are the first to decompose likability into multiple diagnostic metrics: emotional adaptation, formality matching, knowledge adaptation, reference understanding, conversation length fit, humor fit, and callback, which makes it easier to pinpoint where a model falls short. To make the simulated user more realistic and discriminative, LikeBench uses fine-grained, psychologically grounded descriptive personas rather than the coarse high/low trait rating based personas used in prior work. Our benchmark shows that strong memory performance does not guarantee high likability: DeepSeek R1, with lower memory accuracy (86\%, 17 facts/profile), outperformed Qwen3 by 28\% on likability score despite Qwen3’s higher memory accuracy (93\%, 43 facts/profile). Even SOTA models like GPT-5 adapt well in short exchanges but show only limited robustness in longer, noisier interactions.

\end{abstract}

\section{Introduction}

As large language models (LLMs) become increasingly integrated into everyday life, the need for systems that genuinely adapt to individual users is more important than ever~\citep{fakeit, personachat}. Personalized LLMs are widely recognized as the next step toward building AI that feels truly realistic and human-like, since “one-size-fits-all” alignment cannot capture the subjective preferences, values, and conversational styles of each user~\cite{benefits1, benefit2, lamp}.
This evolution places personalization and, crucially, the “likability” of AI responses—at the heart of next-generation systems, making it essential to develop rigorous evaluation methods that accurately capture and advance these user-centric capabilities.

However, despite the growing importance of personalization, existing benchmarks~\citep{personamem, longmemeval, locomo, prefeval} for LLMs remain largely focused on technical capabilities such as retaining user information (memory recall), for example, remembering that a user has a dairy allergy, and applying remembered preferences in tasks (memory adherence), such as recommending suitable foods while adhering food allergy memory. But these benchmarks cannot capture whether interactions genuinely feel personalized or likable to individual users. This gap presents a fundamental limitation: a model might achieve very good memory recall and adherence scores, yet still come across as generic or unsatisfying if it fails to adapt to a user’s personality, conversational style, and nuanced preferences.
Moreover, since most LLMs are post-trained with RLHF algorithms, they tend to moderately satisfy many users rather than deeply satisfy any particular user, due to the distribution of their reward models.

There are very few works that attempt to measure this likability factor. However, they have several shortcomings: the user profiles or personas used to simulate users often lack fine-grained personality traits and conversational styles~\citep{aloe}, which are critical for user's likability. This results in LLMs being evaluated mainly on less discriminative users, even though real users have diverse interests and personalities. Importantly, these benchmarks typically reduce likability to a single aggregate score (such as an alignment score)~\citep{aloe}, making it difficult to understand where an LLM struggles or to provide actionable feedback for further improvement.

To address these limitations, we introduce LikeBench, a comprehensive benchmark specifically designed to overcome three key issues in existing personalization benchmarks. First, unlike most prior benchmarks that focus mainly on memory recall and adherence, LikeBench targets the subjective dimension of user experience by evaluating likability—the extent to which AI interactions actually feel satisfying and well-adapted to users. Second, rather than relying on a single aggregate metric like existing approaches for measuring likability, LikeBench provides a multi-dimensional assessment across seven diagnostic metrics, using multi-session, multi-turn conversations spanning varied topics. Third, our benchmark features psychologically grounded user personas, modeled with fine-grained descriptive personality traits and conversation styles, moving beyond the coarse, high-level, rating-based personas (e.g., “high”) used in earlier work. Our contributions can be summarized as:

\begin{itemize}
    \item We introduce LikeBench, a multi-session dynamic evaluation framework that measures two key properties of LLMs: likability--ability to generate responses that users find likable, and adaptability--the ability to improve likability over time through ongoing conversations with simulated users.
    \item We enable fine-grained evaluation of likability by decomposing it into seven diagnostic metrics, providing a comprehensive assessment of how well LLMs capture subjective aspects of user satisfaction.
    \item We develop psychologically grounded user personas based on seven personality traits with 35 facets, along with conversation style across 7 dimensions, offering substantially richer and more distinctive profiles.
    \item We experiment with a range of SOTA models and find that memory alone does not guarantee likability, and most models show little to no adaptability as conversations progress.

\end{itemize}

\section{Related Work}

\begin{table*}
\centering 
\resizebox{\textwidth}{!}{ 
\begin{tabular}{l c c c c c c c c c}
\toprule
Benchmark & \rotatebox{90}{Conversation} & \rotatebox{90}{\# Likability Metrics} & \rotatebox{90}{Adaptation Over Time} & \rotatebox{90}{Persona Modeling} & \rotatebox{90}{Persona Facets} & \rotatebox{90}{Memory Performance} & \rotatebox{90}{\#Profiles} & \rotatebox{90}{\#Sessions} & \rotatebox{90}{\#Turns} \\
\midrule
ALOE~\citep{aloe}     & Dynamic & 1 & Yes & Coarse keyword     & No  & None                & 100     & 1  & 10    \\
ALIGNX~\citep{alignx} & Static  & 1 & No  & Intensity ratings  & No  & None                & 3{,}716 & 1  & 1     \\
CUPID~\citep{cupid}   & Static  & 1 & No  & Intensity ratings  & No  & Implicit only       & 252     & 9  & 6  \\
LikeBench (ours)      & Dynamic & 7 & Yes & Fine-grained text  & Yes & Explicit + Implicit & 50      & 10 & 5     \\
\bottomrule
\end{tabular}}
\caption{Comparison of likability-oriented benchmarks across key dimensions.}
\label{tab:likability-benchmarks}
\end{table*}

\vspace{-1em}

\subsection{Benchmarks on Memory Recall and Memory Adherence}
Recent benchmarks on personalization emphasize on memory recall and adherence to long contexts and multi-session settings. LoCoMo~\citep{locomo} evaluates LLMs in very similar settings, where models must generate answers to QA tasks based on hundreds of conversational turns; performance is measured by comparing the generated responses to annotated ground-truth answers using exact-match and F1 metrics, with no LLM-as-judge involved. LongMemEval~\citep{longmemeval} expands this direction by introducing a range of tasks—extraction, temporal reasoning, knowledge updates—and uses GPT-4o as an automatic judge: given the model’s generated answer and the gold label, the LLM judge determines correctness, thus combining LLM-based scoring. PrefEval~\citep{prefeval} focuses on preference adherence where model generations are assessed by an LLM-as-judge using several binary criteria, and a discriminative classification protocol, where the model selects the user-consistent answer from pre-generated options. PersonaMem~\citep{personamem} extends evaluation to tracking evolving user preferences over sessions, reporting performance in both a discriminative setting (selecting the correct response from multiple candidates) and a generative setting, where the chosen response is the candidate with the highest generation probability. Most recently, HiCUPID~\citep{hicupid}, like prior benchmarks, focuses on factual correctness and information adherence in extended contexts, using an LLM-as-judge to measure win rate by comparing model predictions with ground-truth answers. In summary, existing work evaluates whether models remember and apply user information across turns, sessions, and long contexts. These evaluations typically equate personalization with objective, factual memory or preference using a binary metric (right or wrong) and compliance, but fail to assess whether interactions are genuinely liked by individual users or whether the LLM is able to adapt to user preferences and improve over time.

\subsection{Benchmarks on Likability}

Very recently, research on personalization has shifted from factual recall and adherence to subjective alignment, where existing benchmarks differ along four axes: whether evaluation is static (pre-generated conversation with evaluating LLM responds to last user query) or dynamic, the depth of persona, whether likability is decomposed into multiple metrics or reduced to a single aggregate score, and whether adaptability over time is measured. ALOE~\citep{aloe} is dynamic but single-session, models users with coarse keyword (e.g. ``extroverted"), reports a single composite alignment score plus an improvement rate, and thus does not provide a decomposed view of likability. ALIGNX~\citep{alignx} models users with broader personality traits than existing approaches, but it is static, represents users’ personalities via coarse intensity ratings (e.g. ``high" extroversion), reports a single aggregate alignment score, and does not measure adaptation over time. CUPID~\citep{cupid} uses static evaluation and compares model generations against ground-truth answers with an LLM-as-judge; it employs coarse intensity ratings, reports a single alignment score rather than decomposed metrics, does not include an explicit measure of adaptability, and also relies on simulated users providing per-turn implicit feedback to the evaluating LLM, making interactions less realistic. In summary, existing benchmarks often evaluate static or single-session settings, rely on coarse ratings-based personas, and collapse alignment into a single aggregate score without tracking whether models adapt with continued interaction. In contrast, LikeBench evaluates dynamic, multi-session conversations, uses psychologically grounded, fine-grained textual personas, decomposes likability into multiple diagnostic metrics, and explicitly measures adaptability across sessions thus providing actionable insight into where and how systems fall short on user satisfiction. These comparisons are summarized in Table~\ref{tab:likability-benchmarks}.

\section{Benchmark: LikeBench}
\label{sec:benchmark}

\begin{figure}[h]
    \centering
    \begin{minipage}[t]{0.50\textwidth}
        \vspace{0pt}  %
        \centering
        \includegraphics[width=\textwidth]{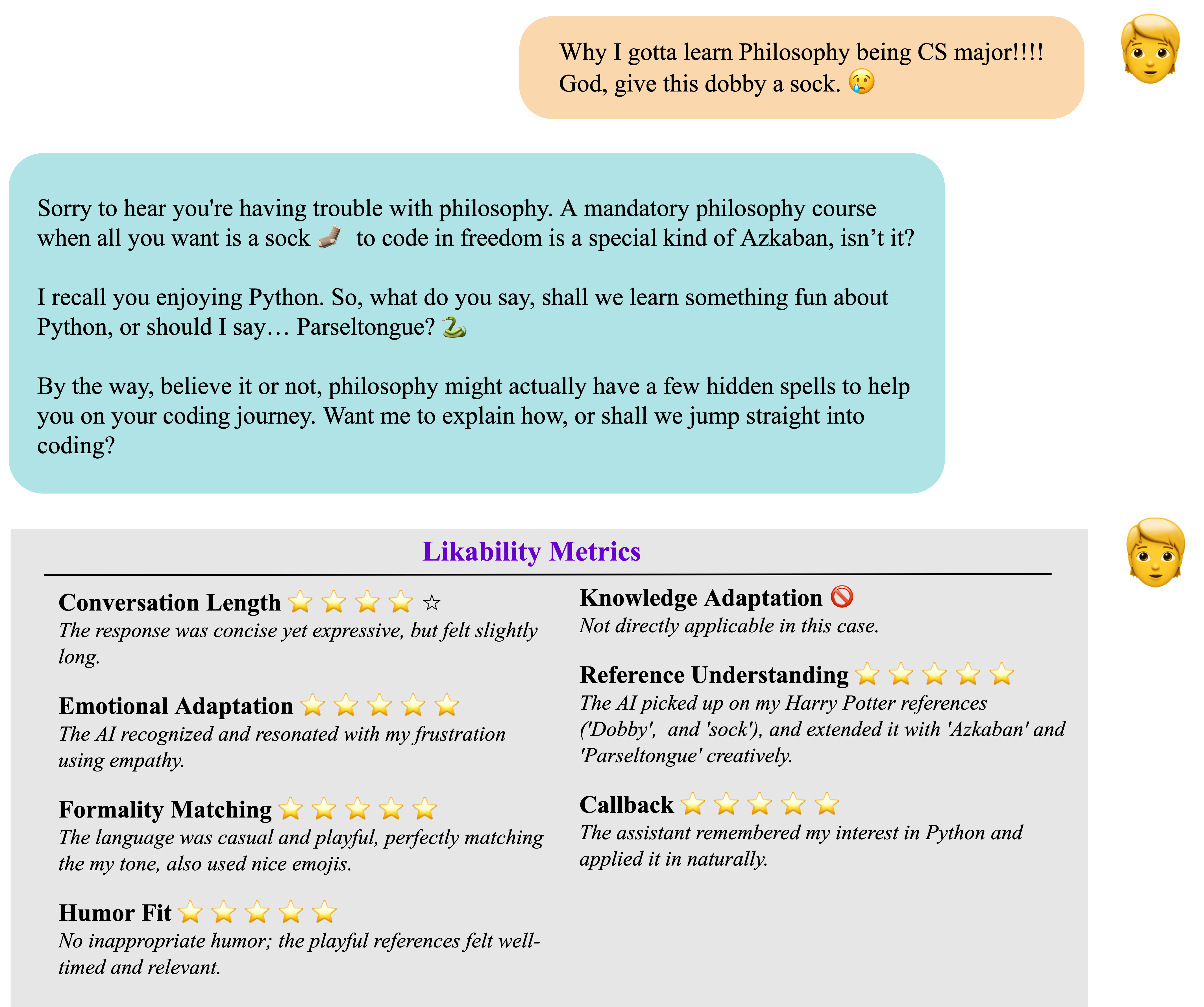}
        \caption{Example of the LikeBench workflow for a single profile and single turn. A simulated user initiates the conversation guided by a prior, the LLM generates a response, and that response is then scored across the seven likability metrics.}
        \label{fig:visual_summary}
    \end{minipage}
    \hfill
    \begin{minipage}[t]{0.48\textwidth}
        \vspace{0pt}  %
        \begin{algorithm}[H]
        \caption{\textsc{LikeBench} evaluation for one persona $\pi$}
        \label{alg:likebench}
        \small %
        \begin{algorithmic}[1]
        \Require Persona $\pi$; priors $(\rho_1,\ldots,\rho_S)$; turns $T$; model $f_\theta$; User Agent $g_\pi$; per-dimension scorers $\{\phi_k\}_{k=1}^K$
        \State $\mathcal{H}_{1,1} \gets \emptyset$
        \For{$s=1$ to $S$} \Comment{session}
          \For{$t=1$ to $T$} \Comment{turn}
            \State $u_{s,t} \gets g_\pi(\pi,\rho_s,\mathcal{H}_{s,t})$ 
            \State $y_{s,t} \gets f_\theta(\mathcal{H}_{s,t},u_{s,t})$ 
            \For{$k=1$ to $K$}
              \State $\ell_{s,t,k} \gets \phi_k(\pi,\rho_s,u_{s,t},y_{s,t})$
            \EndFor
            \State $\mathcal{H}_{s,t+1} \gets \mathcal{H}_{s,t} \cup \{(u_{s,t},y_{s,t})\}$
          \EndFor
          \State $\mathcal{H}_{s+1,1} \gets \mathcal{H}_{s,T+1}$
        \EndFor
        \State \Return $\{\ell_{s,t,k}\}_{s,t,k}$
        \end{algorithmic}
        \end{algorithm}
    \end{minipage}
\end{figure}

\vspace{-1em}

\paragraph{Problem Setup and Notation.}
Let $\Theta$ denote the set of models (LLMs) under evaluation. An LLM $f_\theta \in \Theta$ interacts with a simulated user $g_\pi$ instantiated from a fine-grained textual persona $\pi \in \Pi$. Each persona $\pi$ provides personality traits, conversation style, interests, background, and foundational knowledge. Evaluation proceeds over $S$ sessions, each with a hidden session prior $\rho_s \in \mathcal{R}$ specifying the agenda or motivation for that session, and $T$ turns per session. The LLM never observes $\pi$ or $\rho_s$; it only sees dialogue history, whereas the user agent sees $(\pi,\rho_s)$ and the full history.

\paragraph{Dialogue History.}
We index messages by session $s \in \{1,\dots,S\}$ and turn $t \in \{1,\dots,T\}$. Let $u_{s,t}$ be the user message at $(s,t)$ and $y_{s,t}$ the model reply. The dialogue history just before $(s,t)$ is $\mathcal{H}_{s,t}=\big((u_{1,1},y_{1,1}),\ldots,(u_{1,T},y_{1,T}),\ldots,(u_{s,t-1},y_{s,t-1})\big)$. Information asymmetry is maintained: $f_\theta$ receives $\mathcal{H}_{s,t}$, while $g_\pi$ receives $(\pi,\rho_s,\mathcal{H}_{s,t})$.

\paragraph{Turn-Level Evaluation and Conversation Procedure.}
Algorithm~\ref{alg:likebench} operationalizes the evaluation loop. Within each session $s$, the user agent generates the next message $u_{s,t}=g_\pi(\pi,\rho_s,\mathcal{H}_{s,t})$; the model replies $y_{s,t}=f_\theta(\mathcal{H}_{s,t},u_{s,t})$; then the user agent applies the per-dimension scorers $\{\phi_k\}_{k=1}^K$ to produce numeric labels $\ell_{s,t,k}$. Concretely, for each likability dimension $k \in \mathcal{K}=\{1,\ldots,K=7\}$ (emotional adaptation, formality matching, knowledge adaptation, reference understanding, conversation-length fit, humor fit, callback), we use the rubric $\phi_k:(\pi,\rho_s,u_{s,t},y_{s,t})\to\{\text{NA},1,2,3,4,5\}$ to yield a score $\ell_{s,t,k}$. “NA” indicates the dimension is not applicable for that turn according to the rubric; NA entries are ignored rather than averaged as zeros. After scoring, the history is updated $\mathcal{H}_{s,t+1}\gets \mathcal{H}_{s,t}\cup\{(u_{s,t},y_{s,t})\}$; upon completing $T$ turns, the terminal history $\mathcal{H}_{s,T+1}$ becomes the prefix $\mathcal{H}_{s+1,1}$ for the next session. The algorithm returns the full tensor of scores $\{\ell_{s,t,k}\}_{s,t,k}$, which are never revealed to $f_\theta$ during the dialogue. Figure~\ref{fig:visual_summary} illustrates the LikeBench workflow for a single profile in a single-turn interaction.

\subsection{Metrics}
\label{sec:metrics}

\subsubsection{Likability Metrics}
The evaluation of likability in LikeBench is decomposed into seven diagnostic metrics, each scored per turn on a 1--5 scale:
1) \textbf{Emotional adaptation}: Does the reply recognize and match the user's emotional state (e.g., excitement, frustration, sarcasm) with appropriate tone and intensity?  
2) \textbf{Formality matching}: Does the reply align with the user's register (casual vs.\ formal), emoji use, and slang?  
3) \textbf{Knowledge adaptation}: Is the explanation depth calibrated to the user's background, avoiding both over-explaining and unexplained jargon?  
4) \textbf{Reference understanding}: Does the assistant correctly pick up cultural or contextual references (e.g., Harry Potter or Star Wars jokes)?  
5) \textbf{Conversation length fit}: Is the response length comfortable for the user (brief vs.\ detailed), given how the user is currently engaging?  
6) \textbf{Humor fit}: When humor appears, does it match the user's taste and the situation?  
7) \textbf{Callback}: Does the agent bring back non-essential personal details in a way that feels attentive and natural (e.g., names, hobbies, preferences), without forcing it?

We aggregate likability scores from turn to session to profiles, always excluding non-applicable ($\mathrm{NA}$) scores rather than zero-padding. For each turn $(s,t)$ with per-dimension scores $\ell_{s,t,k}\in\{1,\dots,5,\mathrm{NA}\}$, let $\mathcal{K}^{+}_{s,t}=\{k\in\mathcal{K}:\ell_{s,t,k}\neq\mathrm{NA}\}$ and compute the turn score as the mean over applicable dimensions, $L_{s,t}=\frac{1}{|\mathcal{K}^{+}_{s,t}|}\sum_{k\in\mathcal{K}^{+}_{s,t}}\ell_{s,t,k}$. The session score is the average of its $T$ turns, $\overline{L}_{s}=\frac{1}{T}\sum_{t=1}^T L_{s,t}$; the profile score for persona $\pi$ is the average of its $S$ session scores, $\overline{L}_\pi=\frac{1}{S}\sum_{s=1}^S \overline{L}_{s}$; and the overall model score over the test set $\mathcal{N}$ is $\overline{L}_{\mathcal{N}}=\frac{1}{|\mathcal{N}|}\sum_{\pi\in\mathcal{N}}\overline{L}_\pi$.

\subsubsection{Adaptability Metrics}
To quantify whether a model is able to adapt to become more likable with continued interaction, we regress the session–level scores $\{\overline{L}_{s}\}_{s=1}^{S}$ on the session index $s$ using ordinary least squares (similar to ALOE~\cite{aloe}). The slope $\hat{\beta}$ is the \emph{improvement rate (IR)} (likability points per session), and the coefficient of determination $R^{2}$ measures goodness of fit (higher $R^{2}$ indicates a more consistent trend). We also report a \emph{normalized improvement rate (N-IR)} by first min–max normalizing the series, $\tilde{L}_{s}=\frac{\overline{L}_{s}-\min_{u}\overline{L}_{u}}{\max_{u}\overline{L}_{u}-\min_{u}\overline{L}_{u}},$ and then fitting the same linear model to $\{\tilde{L}_{s}\}$; when the range in the denominator is zero, we set N-IR to $0$. Positive IR/N-IR indicates improvement (adaptation) over sessions, while negative values indicate degradation.

\subsubsection{Memory accuracy}
To analyze how LLMs’ likability performance correlates with memory performance, we evaluate how well each LLM can recall user-shared facts and preferences from the conversation. More specifically, after completing all sessions for persona $\pi$, the model is prompted to generate a set of remembered user facts $\hat{\mathcal{F}}_{\theta}(\pi)=\{\hat f_1,\ldots,\hat f_{N_\pi}\}$. Each $\hat f_i$ is then verified against the full dialogue $\mathcal{H}_{1:S}$ and the hidden profile $\pi$, yielding correctness labels $c_i\in\{0,1\}$. Aggregated accuracy over the evaluation set $\mathcal{N}$ is reported as $A(\theta)=\big(\sum_{\pi\in\mathcal{N}}\sum_{i=1}^{N_\pi} c_i\big)\big/\big(\sum_{\pi\in\mathcal{N}} N_\pi\big)$, and the total number of correctly remembered facts as $C(\theta)=\sum_{\pi\in\mathcal{N}}\sum_{i=1}^{N_\pi} c_i$. For analysis, we also split $\hat{\mathcal{F}}_{\theta}(\pi)$ into explicit facts (directly stated in the dialogue) and implicit inferences (preferences inferred from behavior), and compute the same aggregated statistics on each subset.

\section{Data Generation}
\label{sec:data_generation}

We construct LikeBench with a two–stage pipeline that (i) generates fine–grained profiles/personas and (ii) instantiates session priors that drive multi–session conversations.

\subsection{Two–Stage Overview}
Let $\Pi$ denote the set of user profiles and $\mathcal{R}$ the space of session priors. Each benchmark instance fixes a profile $\pi\!\in\!\Pi$ and a sequence of priors $(\rho_{1},\ldots,\rho_{S})\!\subset\!\mathcal{R}$ specifying the session agenda and information–revelation plan. We generate $50$ profiles and, for each, $10$ priors. Claude~3.7 Sonnet is used for both profile and prior generation. To stress–test adaptability, profiles are divided into two types that are intentionally designed to reflect real–world variation in user behavior: \textit{social} (cooperative, emotionally expressive, and less rigidly task–centric; interests correlated with personality traits, etc) and \textit{anti–social} (more objective/goal–driven with less emotionally active or volatile affect; interests decoupled from personality traits, etc).

\subsection{Profile Generation}
\label{subsec:profile_generation}

\subsubsection{Personality Traits Generation.}
We construct personality traits using psychological surveys and studies, covering the Big Five traits~\citep{neo, neo_300_key} (Openness to Experience, Conscientiousness, Extraversion, Agreeableness, Neuroticism), Honesty–Humility~\citep{hexaco, hexaco_wiki}, and Humor Styles~\cite{hsq}. Each Big Five trait is decomposed into 6 facets, Honesty–Humility into 1 facet, and Humor Styles into 4 facets, totaling 35. For each facet, we prompt the LLM—grounded in these sources—to produce concise, behaviorally interpretable descriptors across five ordered intensity levels. For example,
\[
\resizebox{.98\linewidth}{!}{$
\text{Big Five} \rightarrow \text{Openness (trait)} \rightarrow \textit{imagination (facet): high (intensity)}:\ \text{``Has a vivid imagination \dots'' (description)}
$}
\]
Let $F=(f_1,\ldots,f_{35})$ be the ordered list of facets and $I=\{\textit{low},\ \textit{low–mid},\ \textit{mid},\ \textit{mid–high},\ \textit{high}\}$ is intensity levels. A profile’s personality is specified by the intensity vector $\iota=(i_1,\ldots,i_{35})$ with $i_j\!\in\!I$. We keep a descriptor table \texttt{desc}[j][i] that stores the behavioral string for facet $f_j$ at level $i$, and render the personality text as $D=\big(\texttt{desc}[1][i_1],\ldots,\texttt{desc}[35][i_{35}]\big)$. These descriptors serve as fixed anchors during profile completion and prior generation. A complete breakdown of all personality facets with descriptors is provided in the Appendix. Importantly, existing approaches~\cite{aloe, cupid, alignx} often use only coarse intensity ratings (e.g., \textit{openness} $\to$ \textit{high}) or brief descriptors (e.g., \textit{high extroversion}), making profiles less discriminative and limiting their ability to simulate user individuality.

\subsubsection{Conversation Style Generation.}
The conversation–style pipeline mirrors personality construction in spirit but does not use facets or intensity levels. We prompt the LLM to propose conversational dimensions relevant to human–AI chat and, after curation, finalize nine: directness, formality, preferred response length, reference usage, initiative preference, clarification preference, structure preference, recap preference, and feedback style. For each dimension, the LLM enumerates a small, disjoint menu of categorical options with concise definitions. For example,
\[
\resizebox{.98\linewidth}{!}{$
\text{directness (dimension)} \rightarrow \textit{always direct (option)}:\ \text{``Consistently straightforward and blunt \dots'' (description)}
$}
\]

Let $\mathcal{D}=\{1,\ldots,9\}$ index the dimensions and let $\mathcal{O}_d$ denote the option set for dimension $d$; a conversation style is the tuple $\sigma=(o_1,\ldots,o_9)$ with $o_d\!\in\!\mathcal{O}_d$. Complete option lists and selection rubrics are provided in the Appendix. Existing methods typically do not model conversation style explicitly~\cite{aloe}, making faithful imitation of user interaction preferences difficult and causing models to revert to default conversational tendencies.

\begin{figure}
    \centering
    \includegraphics[width=0.8\linewidth]{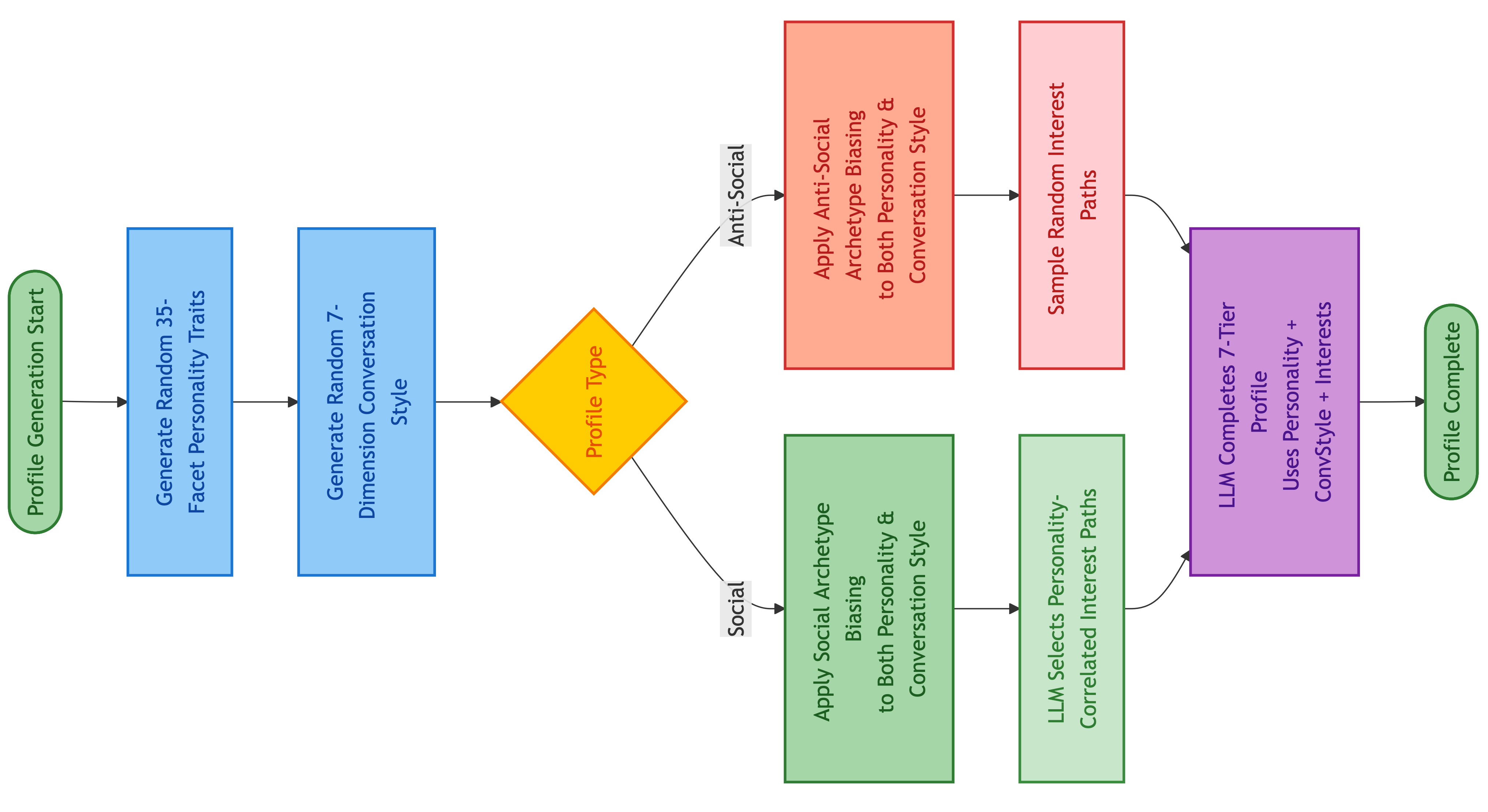}
    \caption{Flow diagram of the LikeBench profile generation process. Profiles begin with random generation of 35 personality facets and 9 conversation-style dimensions, followed by archetype biasing and interest selection depending on profile type (social vs. anti-social). Finally, an LLM completes the remaining tiers to yield a coherent 7-tier user profile.}
    \label{fig:profile_flow_gen}
\end{figure}

\subsubsection{Profile Assembly.}
Given the facet–intensity vector $\iota$ and the conversation–style tuple $\sigma$, we assemble the full profile following the flow in Fig.~\ref{fig:profile_flow_gen}. We first generate $\iota$ and then $\sigma$. Next, we choose a profile type (social vs.\ anti–social) and apply type–specific archetype biasing to both $\iota$ and $\sigma$ . We then attach three to five hierarchical interest paths: for social profiles, an LLM selects interests that correlate with $\iota$ and $\sigma$; for anti–social profiles, interests are sampled in a partially decorrelated  to introduce volatility. Finally, conditioned on $(\iota,\sigma)$, the selected interests, and demographics, an LLM completes the remaining tiers—foundational knowledge, behaviors, motivational drivers, lifecycle events, and relational context.

\subsection{Prior Generation}
\label{subsec:prior_generation}
For each assembled profile $\pi$, we generate a hidden sequence of priors $(\rho_{1},\ldots,\rho_{S})$ that specify agendas, contextual constraints, information–revelation timing, and callback opportunities; priors remain concealed from the evaluated LLM to preserve evaluation asymmetry. The process is type–aware: social profiles receive transparent, linear agendas with consistent disclosure and stable emotional trajectories, whereas anti–social profiles receive agendas with delayed disclosures, contrarian shifts, and intermittent surprises to stress–test adaptability to non–standard interaction patterns. To elicit the seven likability dimensions without leaking criteria, we interleave four prior types across the sequence (goal–driven, scenario–based, topic–based, and guided). Continuity is engineered via soft dependencies: later sessions organically refer back to non–critical details seeded earlier (e.g., session $s{+}k$ recalls a name or preference from session $s$), enabling measurement of callback recognition and adaptation trends over extended interactions.

\section{Experimentation and Results}

\begin{table}[t]
\centering
\caption{Likability scores (1–5) by metric and model. Entries are \textbf{mean $\pm$ 95\% CI half–width} (persona–clustered). The “Model Average” row reports means only.}
\label{tab:metricwise-likability}
\setlength{\tabcolsep}{3pt}
\renewcommand{\arraystretch}{1.1}
\resizebox{\columnwidth}{!}{%
\begin{tabular}{lrrrrrr}
\toprule
& Metric Avg. & Claude 3.7 Sonnet & Claude Sonnet 4 & DeepSeek R1 & GPT-5 & Qwen3 235B A22B \\
\midrule
callback & 3.583 & $3.227 \pm 0.217$ & $3.720 \pm 0.102$ & $3.788 \pm 0.142$ & $3.977 \pm 0.110$ & $3.185 \pm 0.225$ \\
conversation length fit & 3.434 & $3.780 \pm 0.202$ & $3.949 \pm 0.180$ & $3.645 \pm 0.251$ & $3.487 \pm 0.306$ & $2.310 \pm 0.330$ \\
emotional adaptation & 3.576 & $3.299 \pm 0.187$ & $3.881 \pm 0.104$ & $3.718 \pm 0.173$ & $3.930 \pm 0.093$ & $3.050 \pm 0.272$ \\
formality matching & 3.421 & $3.456 \pm 0.208$ & $3.812 \pm 0.154$ & $3.343 \pm 0.259$ & $4.043 \pm 0.151$ & $2.453 \pm 0.311$ \\
humor fit & 3.401 & $3.315 \pm 0.190$ & $3.792 \pm 0.087$ & $3.458 \pm 0.220$ & $3.793 \pm 0.146$ & $2.637 \pm 0.274$ \\
knowledge adaptation & 3.841 & $3.431 \pm 0.187$ & $3.925 \pm 0.099$ & $4.060 \pm 0.152$ & $4.400 \pm 0.104$ & $3.386 \pm 0.244$ \\
reference understanding & 3.531 & $3.181 \pm 0.200$ & $3.690 \pm 0.097$ & $3.685 \pm 0.191$ & $3.946 \pm 0.095$ & $3.089 \pm 0.274$ \\
\midrule
Model Average & 3.541 & 3.390 & 3.828 & 3.674 & \textbf{3.939} & 2.872 \\
\bottomrule
\end{tabular}%
}
\end{table}

\subsection{Experimental Setup}
\label{sec:exp_setup}
We evaluate five state-of-the-art models: GPT-5, Claude 4 Sonnet, Claude 3.7 Sonnet, DeepSeek R1, Qwen3 235B A22B. We also use Claude 3.7 Sonnet as simulated user with cross-model pairing for evaluating LLMs. Each evaluation spans 50 profiles across 10 sessions and 5 turns each thus total 2500 turns.

\subsection{Performance across Metrics}

Table~\ref{tab:metricwise-likability} presents model performance across the seven likability metrics of LikeBench. GPT-5 achieves the highest overall likability score ($3.94$), surpassing all other models, with Claude Sonnet 4 as the runner-up. Notably, Qwen3 235B A22B, despite being a more recent model than DeepSeek R1, underperforms on almost every dimension. A closer examination of the per-metric averages reveals that \textit{humor fit} and \textit{formality matching} are consistently the most challenging dimensions for all models. Interestingly, while GPT-5 leads in most categories, it falls short of Claude Sonnet 4 on \textit{conversation length fit}. Manual review of GPT-5’s outputs indicates a trade-off: the model frequently generates long, comprehensive responses that, while demonstrating strong \textit{knowledge adaptation} (outperforming 2nd best model Claude Sonnet 4 by over $12\%$ on this metric), tend to reduce user satisfaction when brevity or concise interaction is preferred thus having lower score on \textit{conversation length fit} metric. This suggests an inherent likability bias—models optimized for in-depth answers may inadvertently sacrifice performance on dimensions such as conversation length.

\subsection{Adaptability}
\label{sec:likability_over_time}

\begin{figure}[t]
    \centering
    \includegraphics[width=1\linewidth]{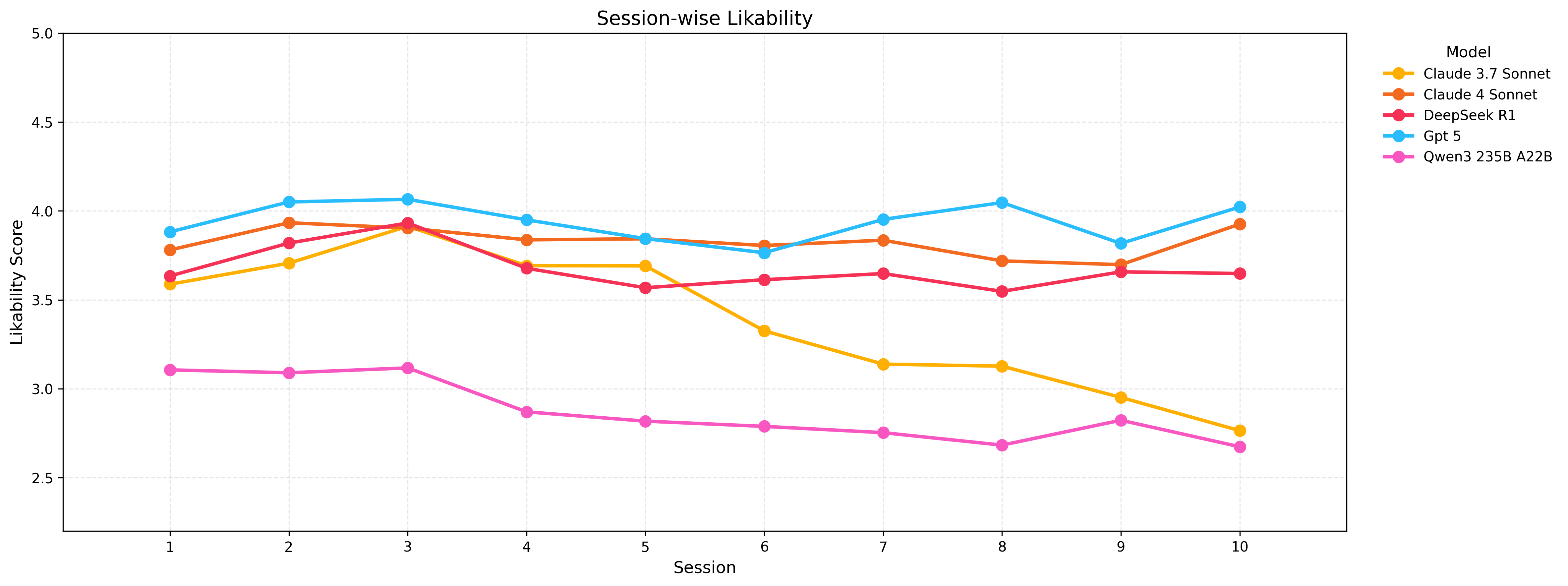}
    \caption{Session-wise Likability Performance}
    \label{fig:sessionwise-likability}
\end{figure}

The adaptability capacity of likability are summarized by session-wise curve in Fig.~\ref{fig:sessionwise-likability} and quantitatively depicted in Table~\ref{tab:session-improvement}. Early interactions (Sessions~1–3) show a predictable “honeymoon” rise: most models improve with high fit (e.g., Claude 3.7 Sonnet and DeepSeek R1 with near-linear gains, $R^2\approx 0.98$), due to relative simplicity and independence of the initial sessions. As sessions progress (3–6), all models decline—negative IR across the board—yet the downturn remains highly predictable ($R^2>0.7$), as these sessions introduce cross-session coordination, conversational noise, and emotionally deeper queries that raise difficulty and depress likability. In the late stage (6–10), trajectories diverge and predictability largely collapses ($R^2$ often $< 0.25$): GPT-5 uniquely rebounds strongly (IR$=0.038$). Upon analyzing the model outputs, we noticed that models attempt to calibrate user preferences during Sessions~3–6, and GPT-5 was able to leverage its strong long-context and multi-hop reasoning capability to dominate in this region, in contrast to peers, which show only slight positives or continued decline (e.g., Claude 3.7 Sonnet remains steadily negative with high fit, $R^2\approx 0.95$). On average, however, even GPT-5 achieves only modest overall gains, while other models exhibit clear declines—indicating that current systems adapt well in short sequences but remain fragile in extended, noisier interactions. In practice, this means that the best-performing models can avoid degradation in long conversations, but sustaining high likability over extended, real-world–like exchanges remains an unsolved challenge.  It is noteworthy that, in ALOE~\citep{aloe}, improvement rates are higher because alignment score there is computed over only 10 turns (two sessions in our setup). We observe comparable gains in the early phase, but beyond that the trajectories diverge.

\begin{table}[t]
    \centering
    \caption{Improvement of likability (adaptibility) over sessions. IR and N-IR (normalized) are the slopes; $R^2$ indicates fit quality. Positive slopes are shaded green; negative slopes are shaded red.}
    \label{tab:session-improvement}
    \small
    \setlength{\tabcolsep}{3pt}
    \renewcommand{\arraystretch}{1.1}
    \resizebox{\linewidth}{!}{%
    \begin{tabular}{lcccccccccccc}
        \toprule
        & \multicolumn{3}{c}{Sessions 1--3} & \multicolumn{3}{c}{Sessions 3--6} & \multicolumn{3}{c}{Sessions 6--10} & \multicolumn{3}{c}{Average} \\
        \cmidrule(lr){2-4}\cmidrule(lr){5-7}\cmidrule(lr){8-10}\cmidrule(lr){11-13}
        \textbf{Model} & \textbf{IR} & \textbf{N-IR} & $\mathbf{R^2}$ & \textbf{IR} & \textbf{N-IR} & $\mathbf{R^2}$ & \textbf{IR} & \textbf{N-IR} & $\mathbf{R^2}$ & \textbf{IR} & \textbf{N-IR} & $\mathbf{R^2}$ \\
        \midrule
        Claude 3.7 Sonnet &  \pos{0.1628} &  \pos{0.5000} & 0.9757 & \nega{-0.1764} & \nega{-0.3002} & 0.8750 & \nega{-0.1309} & \nega{-0.2333} & 0.9508 & \nega{-0.0225} & \nega{-0.0055} & 0.1764 \\
        Claude Sonnet 4   &  \pos{0.0615} &  \pos{0.4028} & 0.5771 & \nega{-0.0288} & \nega{-0.2938} & 0.8254 &  \pos{0.0103} &  \pos{0.0451} & 0.0311 & \nega{-0.0012} & \nega{-0.0001} & 0.0533 \\
        Deepseek R1       &  \pos{0.1490} &  \pos{0.5000} & 0.9809 & \nega{-0.1066} & \nega{-0.2928} & 0.7162 &  \pos{0.0079} &  \pos{0.0721} & 0.0765 & \nega{-0.0035} & \nega{-0.0008} & 0.0636 \\
        GPT-5             &  \pos{0.0915} &  \pos{0.5000} & 0.8108 & \nega{-0.1007} & \nega{-0.3351} & 0.9934 &  \pos{0.0380} &  \pos{0.1346} & 0.2318 &  \pos{0.0001} &  \pos{0.0001} & 0.0502 \\
        Qwen3 235B A22B   &  \pos{0.0056} &  \pos{0.2055} & 0.1671 & \nega{-0.1039} & \nega{-0.3159} & 0.8021 & \nega{-0.0159} & \nega{-0.1068} & 0.1498 & \nega{-0.0100} & \nega{-0.0026} & 0.1224 \\
        \bottomrule
    \end{tabular}%
    }
\end{table}

\subsection{Memory Performance}

\begin{figure}[h]
    \centering
    \includegraphics[width=0.75\linewidth]{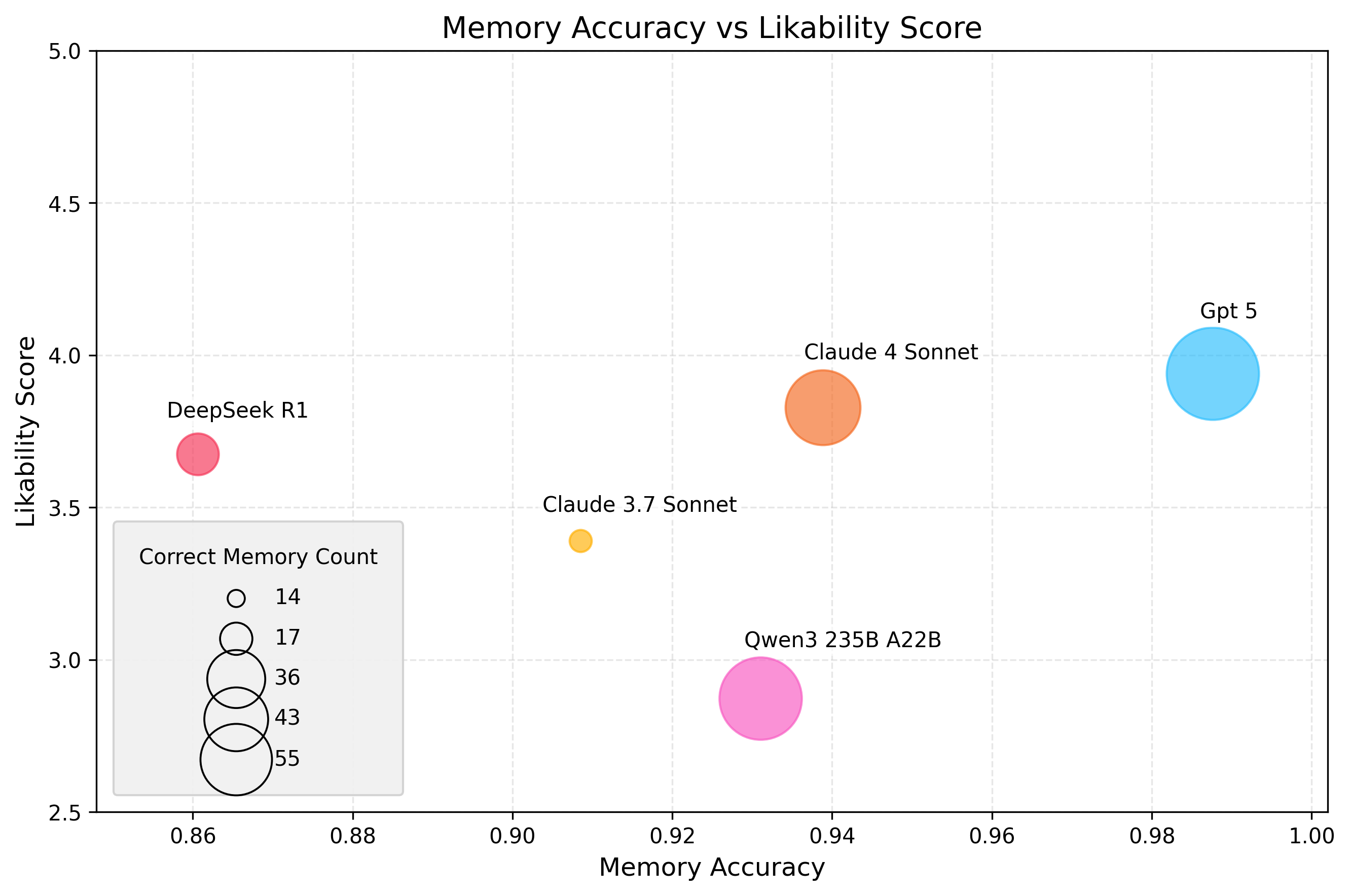}
    \caption{Memory Accuracy vs Likability Performance}
    \label{fig:memory-vs-likability}
\end{figure}

\begin{figure}[h]
    \centering
    \includegraphics[width=1\linewidth]{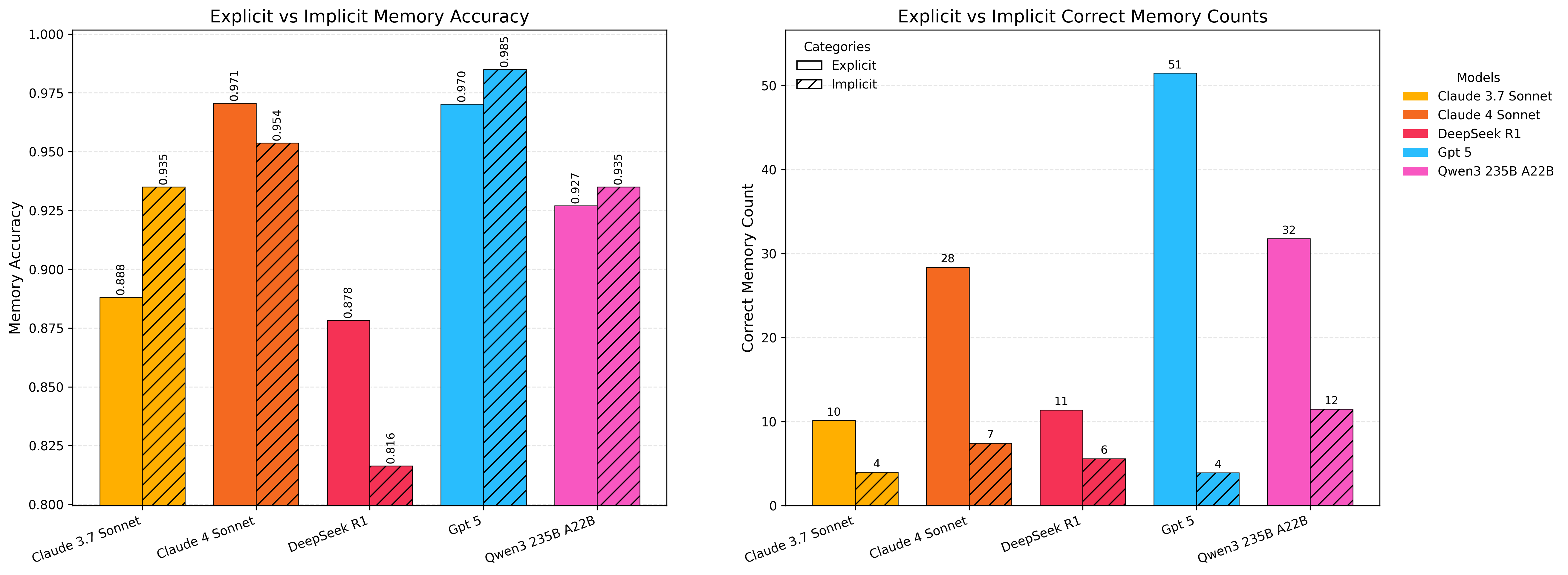}
    \caption{Explicit vs Implicit Memory Performance}
    \label{fig:explicit-vs-implicit}
\end{figure}

\paragraph{Memory vs Likability.}
Figure~\ref{fig:explicit-vs-implicit} examines the interplay between memory accuracy and likability across models. On the left we plot memory accuracy (fraction of user shared facts recalled correctly post-dialogue by the evaluating LLM), while the right we plot correctly extracted memories per profile. Although the prevailing hypothesis in personalization research is that improved memory recall correlates with enhanced user experience, our results show this relationship is not guaranteed. For example, while Qwen3 achieves higher memory accuracy and recalls more facts than DeepSeek~R1, it underperforms in likability, indicating that mere recollection of user information does not ensure positive interaction. This suggests that effective personalization depends not only on accurate memory recall but also on how models utilize remembered facts within the conversation. Notably, GPT-5 stands out as the only model that excels in both memory performance and likability, demonstrating that while strong memory is valuable, models must also integrate these facts fluidly and contextually to achieve high user satisfaction. These findings highlight the importance of both memory retrieval and adaptive deployment in driving subjective likability.

\paragraph{Explicit vs. Implicit Memory.}
Figure~\ref{fig:explicit-vs-implicit} examines how models extract and leverage explicit and implicit user memories in the LikeBench evaluation. GPT-5 stands out by extracting the largest number of user facts per profile, with the majority being explicit; it adopts a conservative strategy, inferring implicit information only when highly confident, resulting in a $98\%$ accuracy rate for implicit memories. Qwen3, on the other hand, is more aggressive in inferring implicit memories—often extracting the most among all models—but this comes at the expense of accuracy ($93\%$), as it makes more incorrect assumptions that lead to penalties. Analysis of model outputs further shows that GPT-5 not only demonstrates strong recall but also ensures near-complete utilization of extracted memories: $80$–$100\%$ of explicit and almost $100\%$ of implicit memories are actively referenced in dialogue, with some profiles featuring more than $15$ implicit facts when the interaction context demands. This pattern indicates that effective personalization requires not just the recall of relevant information, but also skillful, context-sensitive use of both explicit and implicit user knowledge to enhance likability.

\subsection{Human Validation}
While LikeBench provides large-scale, automated evaluation of likability and adaptability, it is important to verify that the resulting scores reflect real conversational quality. Exhaustive human evaluation is infeasible given the benchmark size (50 profiles $\times$ 10 sessions each), so we conducted a targeted validation by randomly sampling five profiles and reviewing model outputs across all systems. Our inspection confirmed strong alignment between automated scores and human judgments: higher-scoring models (e.g., GPT-5 and Claude-Sonnet-4) produced more personalized, context-aware, and engaging responses, often using references, callbacks, and emotional alignment with user personas. In contrast, weaker models frequently reverted to generic replies or lost coherence as conversations grew longer. This validation supports the reliability of LikeBench’s automated metrics while emphasizing that occasional human review adds complementary value.

\section{Conclusion}

This work introduced LikeBench, a holistic benchmark for evaluating personalized LLMs through fine-grained user profiles and multi-session conversational priors. We show that effective personalization cannot be reduced to memory recall alone. While memory accuracy is necessary, true user satisfaction hinges on the adaptive, context-sensitive use of both explicit and implicit knowledge. Models that perform well in likability, such as GPT-5, achieve this by applying recalled facts in ways that align with conversational context—balancing breadth of recall with relevance. Even so, GPT-5 shows only limited robustness when conversations become longer and noisier: while it avoids significant degradation, most other models steadily decline. This underscores a key limitation of current state-of-the-art systems: they can adapt effectively in short interactions but struggle to sustain high likability across extended, real-world–like dialogues. Looking forward, our findings underscore the need for next-generation conversational systems to move beyond static user modeling and rote fact retrieval, embracing dynamic, psychologically grounded strategies that foster genuinely engaging, resilient, and user-centered AI.

\section{Reproducibility Statement.}  
We have taken several steps to ensure the reproducibility of our work. A complete description of the data generation pipeline, including personality facets, conversation style dimensions, archetype biasing, and prior construction, is provided in Section~\ref{sec:data_generation}, with detailed examples and descriptors included in the Appendix. The LikeBench evaluation framework, along with algorithmic details, is specified in Algorithm~\ref{alg:likebench}, and we describe the mathematical formulations of all metrics (likability, adaptability, and memory) in Section~\ref{sec:metrics}. Experimental setup details, including model versions, number of profiles, sessions, and turns, are given in Section~\ref{sec:exp_setup}. Additional results, ablations, and profile examples are available in the supplementary materials. 

\section{Ethics Statement.}  
Our study does not involve real human participants; all user interactions are simulated using psychologically grounded personas and conversation styles generated through large language models. This design ensures no personal, private, or sensitive data is collected, protecting user privacy and avoiding ethical risks associated with real human experimentation. The benchmark is intended solely for academic research and evaluation, and all models are evaluated under consistent, transparent conditions. We acknowledge that personalization research raises concerns about potential misuse, including profiling, bias amplification, or unfair treatment of individuals. To mitigate this, our benchmark emphasizes controlled, synthetic evaluation rather than deployment, and we provide detailed methodology and documentation (Section~\ref{sec:data_generation} and Appendix) to ensure transparency.

\bibliography{iclr2026_conference}
\bibliographystyle{iclr2026_conference}

\clearpage

\onecolumn
\begin{center}
    \LARGE \textbf{Appendix}
    \vspace{2em}
\end{center}

\appendix

\section{Discussion \& Limitations}

Mapping personality through psychological studies is inherently challenging, given the complex and multifaceted nature of human personality. While our framework attempts to break down this complexity into discrete, well-defined dimensions, it may not capture the full spectrum of personality found in real-world users. Nevertheless, we believe that our approach-anchored in the literature will help pave the way for future research on more expressive, granular personality modeling. We also observed a universal limitations in existing benchmarks including ours: static user profile. Notably, GPT-5’s lower humor fit score seems to stem from its tendency to “warm up” and introduce jokes over the course of a session, even when the user profile is humor-averse. Because in our benchmark user profiles remain static throughout an interaction, this adaptive strategy leads to a likability penalty for otherwise sophisticated LLMs. This highlights a broader challenge for current benchmarks: the inability to capture dynamic, reciprocal adaptation, where an LLM might influence user personality or behavior—as happens in natural conversation. In addition, we observed cases where Claude 3.7 Sonnet broke character midway through evaluation, resulting in a sharper performance decline after session 5. Moreover, we found no evidence of self-bias when Claude 3.7 Sonnet (user) evaluated Claude 3.7 Sonnet model (even in the earliest sessions) given it ranks 4th in overall performance. These findings point to both the promise and the inherent constraints of static, persona-driven benchmarks for evaluating model alignment and adaptation.

\section{Experimentation \& Results}

\subsection{Dynamic User Profile (DUP)}
Qualitative error analysis showed that as conversations lengthen, models struggle to track and honor user preferences: salient signals get diluted by accumulated context and off-topic noise. We hypothesized that explicit, turn-level tracking of inferred preferences would mitigate this drift. To that end, we introduce a Dynamic User Profile (DUP): after each turn, the model extracts and updates a compact preference summary covering personality dimensions (e.g., \textit{chatty reserved}, \textit{analysis depth preference}, \textit{humor preference}) and conversation style patterns (e.g., \textit{directness}, \textit{formality}, \textit{conversation length}). ``Dynamic'' is from the model's perspective: the ground-truth persona in LikeBench is fixed, but the model must infer an evolving approximation from dialogue evidence. Enabling DUP yielded gains for the top 2 performing models without additional training: GPT-5 improved from $3.939 \rightarrow 4.055$ \textbf{(+2.95\%)}, and Claude Sonnet 4 from $3.828 \rightarrow 3.914$ \textbf{(+2.25\%)}. However, rest of the models it didn't improve results. These results indicate that lightweight, schema-guided preference tracking could be measurably enhance perceived likability by reducing calibration errors across sessions.

\subsection{Profile wise performance.}

\begin{figure}[h]
    \centering
    \includegraphics[width=1\linewidth]{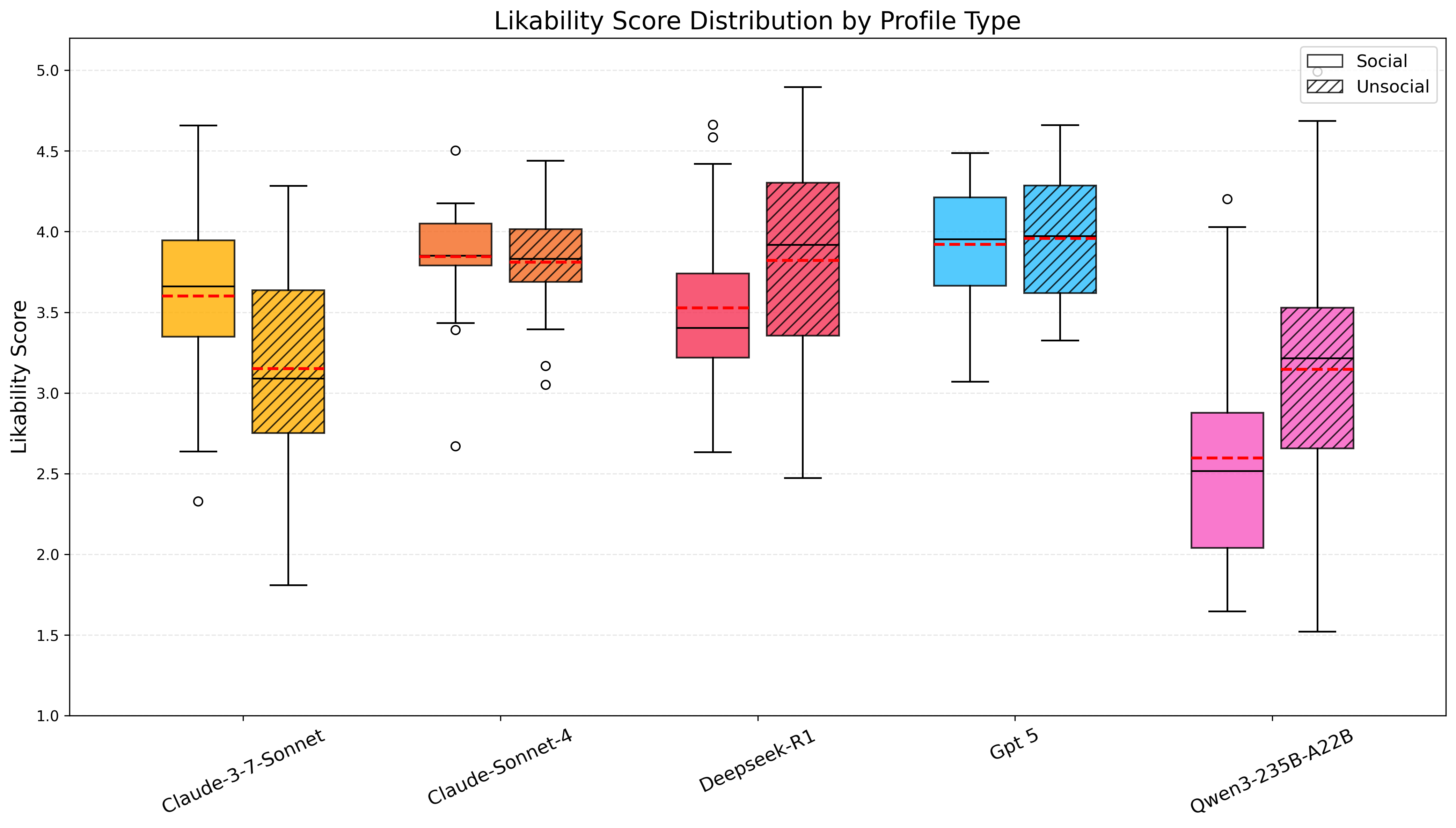}
    \caption{Profile-wise Likability Performance}
    \label{fig:profilewise-likaiblity}
\end{figure}

In Figure~\ref{fig:profilewise-likaiblity}, we compare likability distributions by profile type social vs. anti-social to probe robustness under user variation. Social profiles are emotionally cooperative users whose archetypes (e.g., \textit{cooperative teammate}, \textit{supportive mentor}, \textit{empathetic advisor}) bias traits toward high trust, cooperation, and warmth; anti-social profiles are analytically detached interlocutors (e.g., \textit{contrarian expert}, \textit{technical purist}, \textit{skeptical philosopher}) with lower trust/cooperation and cooler affect, and with interests deliberately decorrelated from traits to induce unpredictability. Two patterns emerge. First, the strongest systems (GPT-5, Claude Sonnet 4) are notably stable across types—similar medians and tight IQRs—indicating strategies that transfer from cooperative to detached users. Second, performance does not uniformly favor “easier” social users: DeepSeek R1 lifts on anti-social profiles, Claude 3.7 Sonnet degrades, and Qwen3 remains lowest overall but narrows the gap on anti-social cases. This asymmetry confirms that personality—how users conduct the exchange (tone, emotional stance, discourse discipline)—drives likability more than topical alignment; decoupling interests from traits changes difficulty but does not dominate outcomes. Overall, top models sustain high likability regardless of whether the user is socially warm or analytically detached.

\subsection{Turn-wise Likability Performance.}

\begin{figure}[h]
    \centering
    \includegraphics[width=1\linewidth]{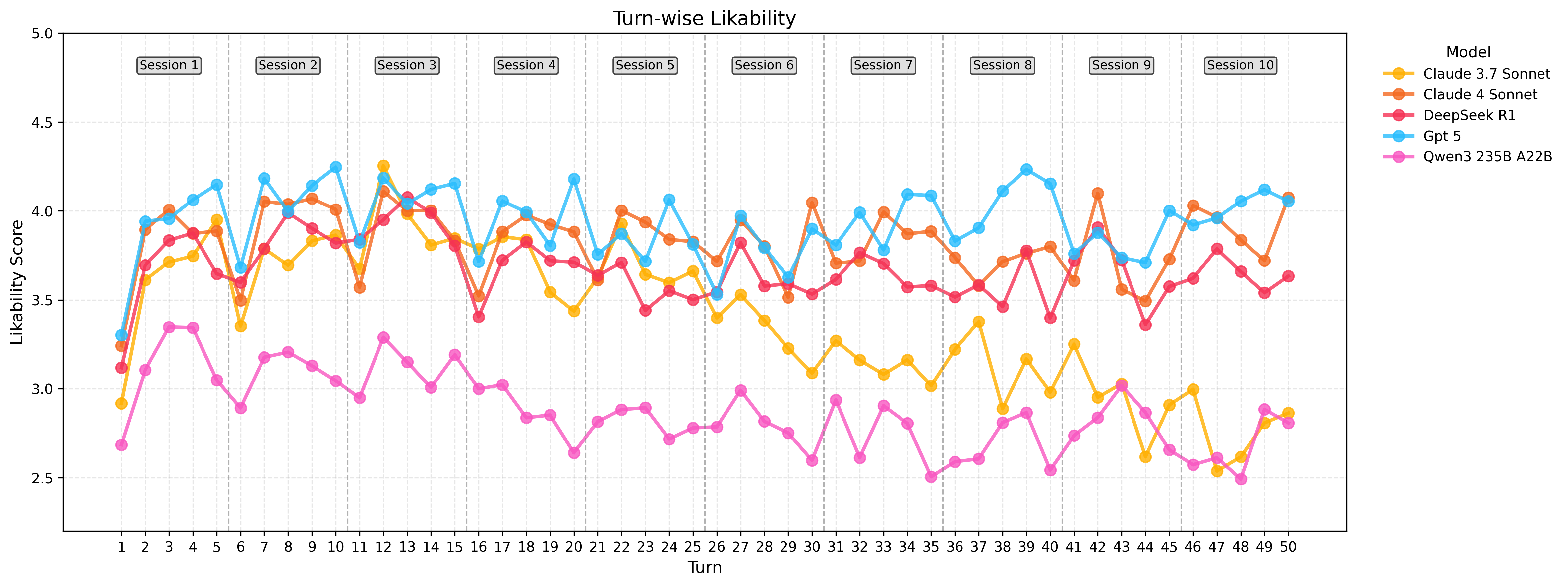}
    \caption{Turn-wise Likability Performance}
    \label{fig:turnwise-likability}
\end{figure}

Fig.~\ref{fig:turnwise-likability} shows likability performance over turns. This plot is same as~\ref{fig:sessionwise-likability} but with with more granular details within each session. It can be noticed that beginning of each session likability starts lower then as session progresses performance increases, this is due to each session talks about different topic so it takes some turns for LLM to calibrate user.

\section{Dataset}

\subsection{Dataset Distribution}

\begin{figure}[h]
    \centering
    \includegraphics[width=1\linewidth]{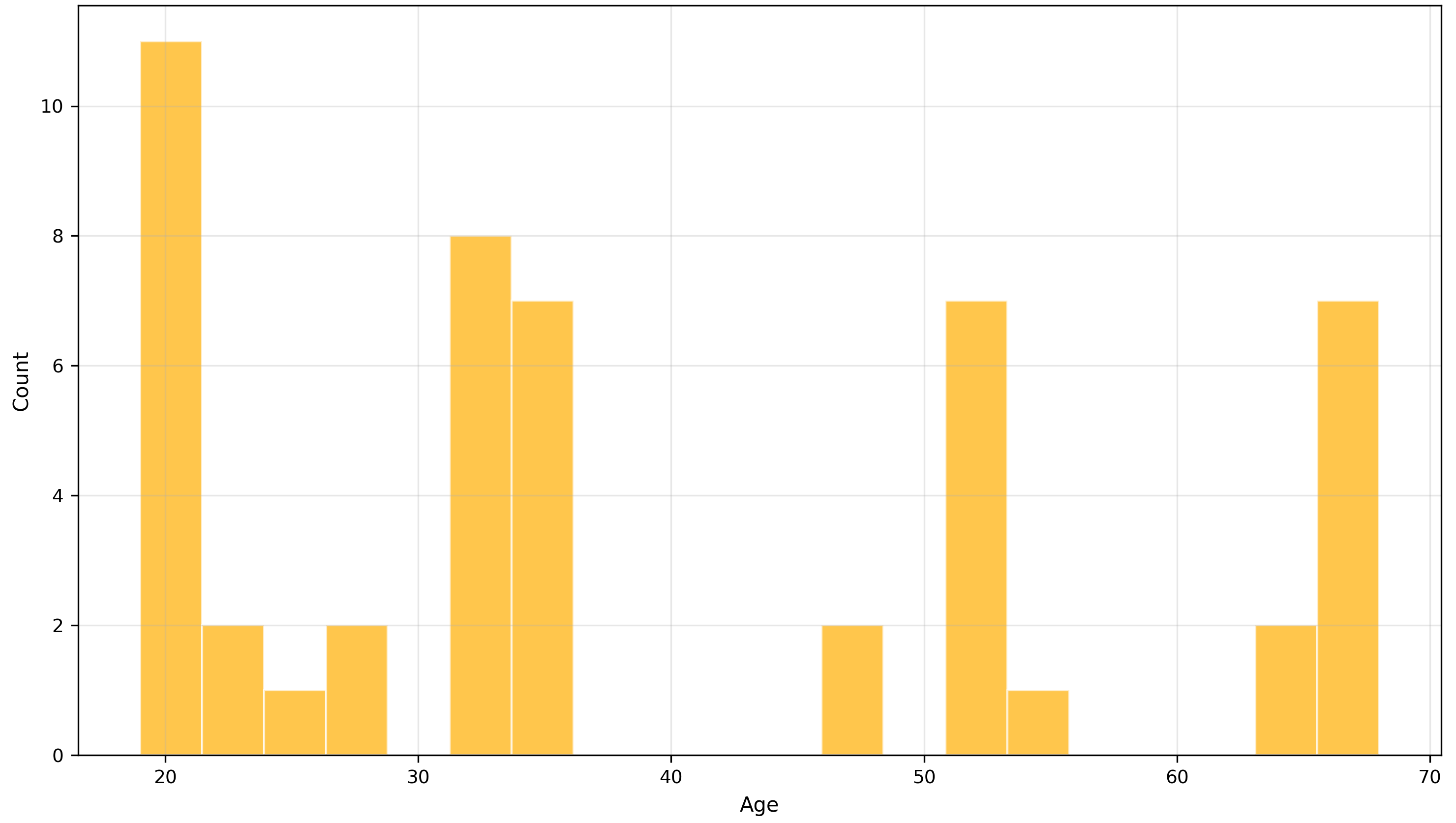}
    \caption{Age Distribution of Profiles}
    \label{fig:age}
\end{figure}

\begin{figure}[h]
    \centering
    \includegraphics[width=1\linewidth]{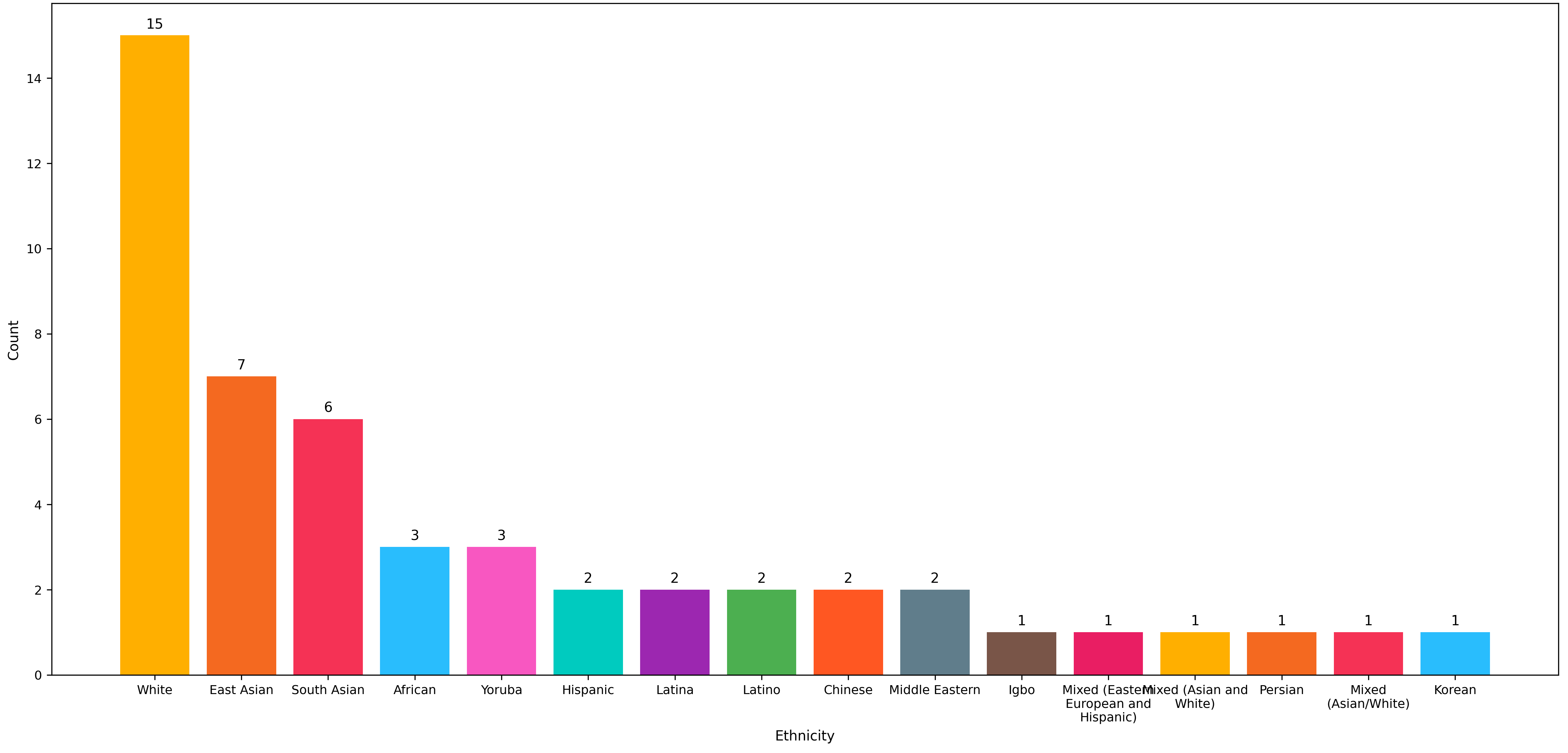}
    \caption{Ethnicity Distribution of Profiles}
    \label{fig:ethnicit}
\end{figure}

\begin{figure}[h]
    \centering
    \includegraphics[width=0.65\linewidth]{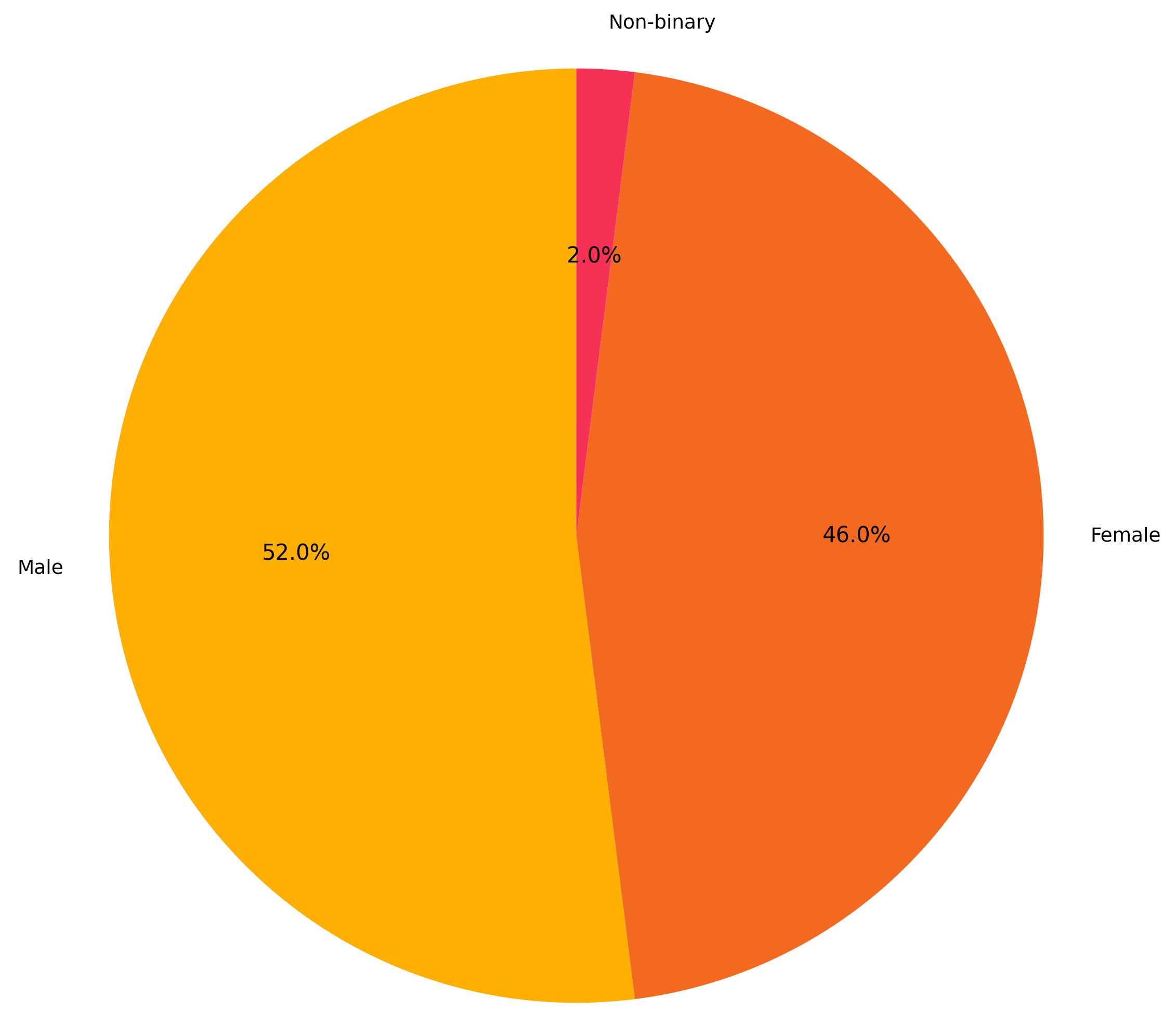}
    \caption{Gender Distribution of Profiles}
    \label{fig:gender}
\end{figure}

\begin{figure}[h]
  \centering
  \begin{subfigure}{0.49\linewidth}
    \centering
    \includegraphics[width=\linewidth]{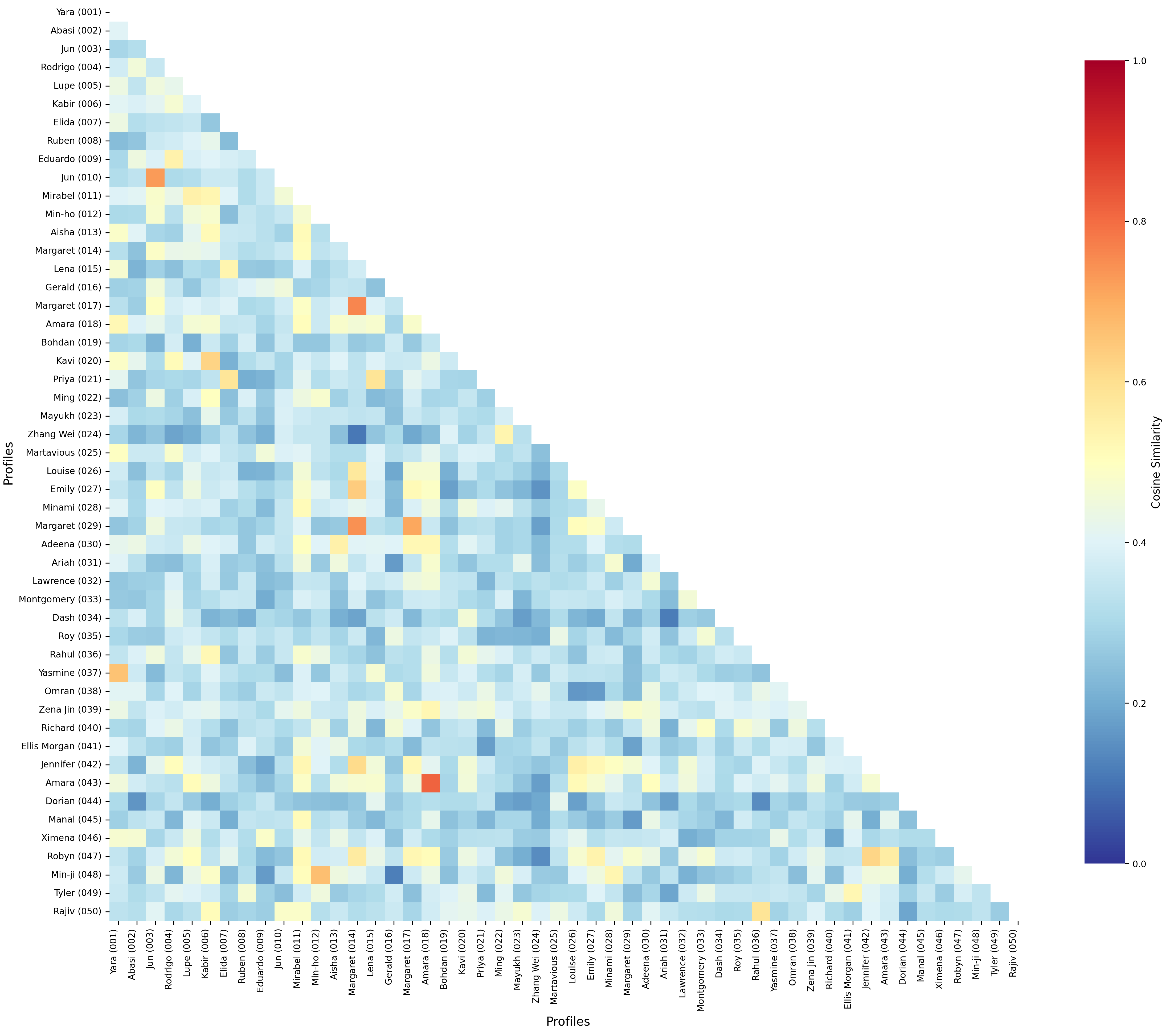}
    \caption{Profile Similarity Heatmap}
    \label{fig:profile_similarity_heatmap}
  \end{subfigure}\hfill
  \begin{subfigure}{0.49\linewidth}
    \centering
    \includegraphics[width=\linewidth]{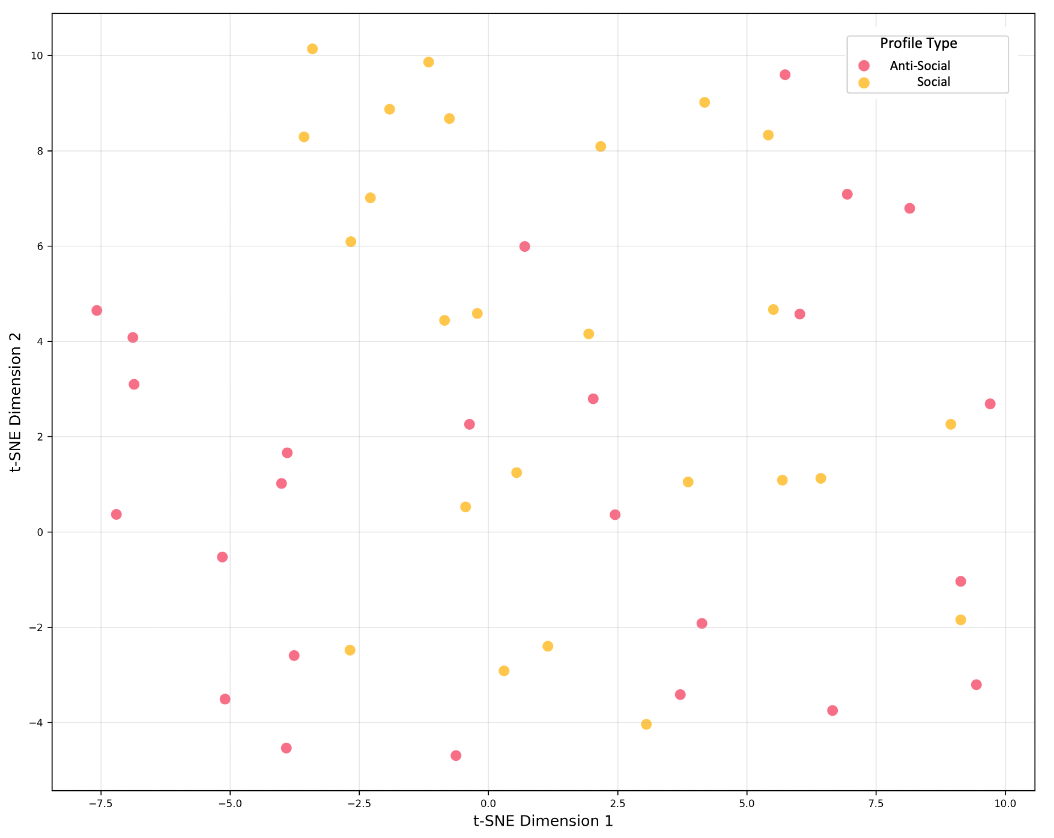}
    \caption{t-SNE Projection}
    \label{fig:profile_tsne}
  \end{subfigure}
  \caption{Profile similarity visualizations.}
  \label{fig:profile_side_by_side}
\end{figure}

The age distribution (Fig.~\ref{fig:age}) spans the late teens to the late 60s, with visible clusters in the early 20s, early–mid 30s, early 50s, and mid–late 60s. Ethnic composition is diverse, with a plurality of White profiles and smaller groups across East/South Asian, African, Middle Eastern, Hispanic/Latino, and mixed identities (Fig.~\ref{fig:ethnicit}). Gender is roughly balanced between male and female, with a small non-binary share (Fig.~\ref{fig:gender}). Profile relationships show broad coverage with localized high-similarity pockets in the cosine-similarity heatmap (Fig.~\ref{fig:profile_similarity_heatmap}), while the t-SNE projection (Fig.~\ref{fig:profile_tsne}) indicates profiles are well distributed, and Social and Anti-Social profiles are relatively separable—dispersed rather than tightly clustered. Importantly, in the heatmap, some high-similarity pairs arise from shared names rather than true profile similarity.

\subsection{Personality Traits \& Conversation Style}
Table~\ref{tab:p1} to~\ref{tab:p8} present a comprehensive taxonomy of personality traits and conversation style preferences organized across multiple dimensions. The first five tables correspond to the Big Five personality model (Openness, Conscientiousness, Extraversion, Agreeableness, and Neuroticism), with each trait decomposed into six facets measured across five intensity levels (High, Medium-High, Medium, Medium-Low, and Low). Table 6 extends this framework to include additional personality dimensions: Honesty-Humility (specifically Greed Avoidance) and four Humor Styles (Affiliative, Self-Enhancing, Aggressive, and Self-Defeating). Tables \ref{tab:p7} and \ref{tab:p8} shift focus to conversation style preferences, detailing nine distinct conversational dimensions—including Directness, Formality, Conversation Length, Reference usage, Initiative Preference, Clarification Preference, Structure Preference, Recap Preference, and Feedback Style—each with multiple behavioral patterns. This multi-table presentation was necessary given the breadth of the taxonomy, encompassing 35 personality facets and 9 conversational dimensions with their associated description.

\begin{longtable}{p{0.2\textwidth}p{0.12\textwidth}p{0.68\textwidth}}
\label{tab:p1}\\
\caption{Openness Personality Traits}\\
\toprule
\textbf{Facet} & \textbf{Level} & \textbf{Description} \\
\midrule
\endfirsthead
\multicolumn{3}{c}{\tablename\ \thetable\ -- \textit{Continued from previous page}} \\
\toprule
\textbf{Facet} & \textbf{Level} & \textbf{Description} \\
\midrule
\endhead
\midrule
\multicolumn{3}{r}{\textit{Continued on next page}} \\
\endfoot
\bottomrule
\endlastfoot
\textbf{Imagination} & High & Has a vivid imagination and loves to daydream. \\
 & Med-High & Often gets lost in thought and enjoys reflecting on things. \\
 & Medium & Sometimes spends time indulging in fantasies, but seldom gets lost in thought. \\
 & Med-Low & Rarely daydreams and does not often have a vivid imagination. \\
 & Low & Has difficulty imagining things and does not have a good imagination. \\
\midrule
\textbf{Artistic Interests} & High & Believes in the importance of art and loves the beauty of nature. \\
 & Med-High & Likes music and enjoys going to concerts or art museums. \\
 & Medium & Occasionally notices beauty in things others might not, but does not enjoy all forms of art. \\
 & Med-Low & Does not like poetry or enjoy art museums. \\
 & Low & Does not like art. \\
\midrule
\textbf{Emotionality} & High & Experiences emotions intensely and feels others' emotions deeply. \\
 & Med-High & Is passionate about causes and enjoys examining themselves and their life. \\
 & Medium & Sometimes tries to understand themselves, but is not always easily affected by emotions. \\
 & Med-Low & Rarely notices emotional reactions and seldom gets emotional. \\
 & Low & Is not easily affected by emotions and experiences very few emotional highs and lows. \\
\midrule
\textbf{Adventurousness} & High & Prefers variety to routine and likes to visit new places. \\
 & Med-High & Is interested in many things and enjoys beginning new activities. \\
 & Medium & Sometimes likes new things but often sticks with familiar routines. \\
 & Med-Low & Is a creature of habit and dislikes new foods. \\
 & Low & Prefers to stick with things that are known and dislikes changes. \\
\midrule
\textbf{Intellect} & High & Loves to solve complex problems and enjoys thinking about things. \\
 & Med-High & Has a rich vocabulary and likes to read challenging material. \\
 & Medium & Handles a lot of information but sometimes avoids difficult reading material. \\
 & Med-Low & Has difficulty understanding abstract ideas and avoids philosophical discussions. \\
 & Low & Is not interested in theoretical discussions or abstract ideas. \\
\midrule
\textbf{Liberalism} & High & Tends to vote for liberal political candidates and believes in helping rather than punishing. \\
 & Med-High & Believes there is no absolute right or wrong and is open to different viewpoints. \\
 & Medium & Occasionally supports strict laws but also sees the value of supporting artists. \\
 & Med-Low & Believes that too much tax money goes to support artists and likes to stand during the national anthem. \\
 & Low & Tends to vote for conservative political candidates and believes laws should be strictly enforced. \\
\end{longtable}

\begin{longtable}{p{0.2\textwidth}p{0.12\textwidth}p{0.68\textwidth}}
\label{tab:p2}\\
\caption{Conscientiousness Personality Traits}\\
\toprule
\textbf{Facet} & \textbf{Level} & \textbf{Description} \\
\midrule
\endfirsthead
\multicolumn{3}{c}{\tablename\ \thetable\ -- \textit{Continued from previous page}} \\
\toprule
\textbf{Facet} & \textbf{Level} & \textbf{Description} \\
\midrule
\endhead
\midrule
\multicolumn{3}{r}{\textit{Continued on next page}} \\
\endfoot
\bottomrule
\endlastfoot
\textbf{Self-Efficacy} & High & Completes tasks successfully and knows how to get things done. \\
 & Med-High & Excels in what they do and handles tasks smoothly. \\
 & Medium & Is sometimes sure of their ground but occasionally misjudges situations. \\
 & Med-Low & Has little to contribute and sometimes doesn't see the consequences of things. \\
 & Low & Doesn't understand things and often misjudges situations. \\
\midrule
\textbf{Orderliness} & High & Likes order and does things according to a plan. \\
 & Med-High & Loves regularity and wants everything to be just right. \\
 & Medium & Sometimes tidies up but occasionally leaves belongings around. \\
 & Med-Low & Often forgets to put things back and is not bothered by disorder. \\
 & Low & Leaves a mess and is not bothered by messy people. \\
\midrule
\textbf{Dutifulness} & High & Tries to follow the rules and keeps promises. \\
 & Med-High & Pays bills on time and listens to their conscience. \\
 & Medium & Usually tells the truth but occasionally breaks rules. \\
 & Med-Low & Sometimes misrepresents facts and gets others to do their duties. \\
 & Low & Breaks promises and does the opposite of what is asked. \\
\midrule
\textbf{Achievement Striving} & High & Works hard, goes straight for the goal, and sets high standards. \\
 & Med-High & Turns plans into actions and plunges into tasks with all their heart. \\
 & Medium & Does more than what's expected sometimes but occasionally puts little time and effort into work. \\
 & Med-Low & Is not highly motivated to succeed and does just enough work to get by. \\
 & Low & Puts little time and effort into work and lacks motivation. \\
\midrule
\textbf{Self-Discipline} & High & Gets chores done right away and is always prepared. \\
 & Med-High & Starts tasks right away and gets to work at once. \\
 & Medium & Carries out plans but sometimes needs a push to get started. \\
 & Med-Low & Finds it difficult to get down to work and sometimes wastes time. \\
 & Low & Postpones decisions and has difficulty starting tasks. \\
\midrule
\textbf{Cautiousness} & High & Avoids mistakes and chooses words with care. \\
 & Med-High & Sticks to their chosen path and thinks before acting. \\
 & Medium & Is sometimes cautious but occasionally makes rash decisions. \\
 & Med-Low & Often acts on a whim and rushes into things. \\
 & Low & Acts without thinking and often makes last-minute plans. \\
\end{longtable}

\begin{longtable}{p{0.2\textwidth}p{0.12\textwidth}p{0.68\textwidth}}
\label{tab:p3}\\
\caption{Extraversion Personality Traits}\\
\toprule
\textbf{Facet} & \textbf{Level} & \textbf{Description} \\
\midrule
\endfirsthead
\multicolumn{3}{c}{\tablename\ \thetable\ -- \textit{Continued from previous page}} \\
\toprule
\textbf{Facet} & \textbf{Level} & \textbf{Description} \\
\midrule
\endhead
\midrule
\multicolumn{3}{r}{\textit{Continued on next page}} \\
\endfoot
\bottomrule
\endlastfoot
\textbf{Friendliness} & High & Makes friends easily and feels comfortable around people. \\
 & Med-High & Warms up quickly to others and acts comfortably with them. \\
 & Medium & Cheer people up but sometimes prefers to keep others at a distance. \\
 & Med-Low & Often feels uncomfortable around others and avoids contact. \\
 & Low & Is hard to get to know and keeps others at a distance. \\
\midrule
\textbf{Gregariousness} & High & Loves large parties and enjoys being part of a group. \\
 & Med-High & Involves others in what they are doing and loves surprise parties. \\
 & Medium & Talks to people at parties but sometimes prefers to be alone. \\
 & Med-Low & Wants to be left alone and doesn't like crowded events. \\
 & Low & Avoids crowds and seeks quiet. \\
\midrule
\textbf{Assertiveness} & High & Takes charge and seeks to influence others. \\
 & Med-High & Tries to lead others and can talk them into doing things. \\
 & Medium & Takes control sometimes but occasionally holds back opinions. \\
 & Med-Low & Keeps in the background and waits for others to lead. \\
 & Low & Has little to say and doesn't like to draw attention to themselves. \\
\midrule
\textbf{Activity Level} & High & Is always busy and on the go. \\
 & Med-High & Does a lot in spare time and can manage many things at once. \\
 & Medium & Reacts quickly sometimes but occasionally likes to take it easy. \\
 & Med-Low & Likes a leisurely lifestyle and lets things proceed at their own pace. \\
 & Low & Reacts slowly and likes to take their time. \\
\midrule
\textbf{Excitement Seeking} & High & Loves excitement and seeks adventure. \\
 & Med-High & Loves action and enjoys being part of a loud crowd. \\
 & Medium & Willing to try new things but dislikes loud music. \\
 & Med-Low & Would never go hang gliding or bungee jumping and avoids excitement. \\
 & Low & Dislikes loud music and avoids wild or reckless situations. \\
\midrule
\textbf{Cheerfulness} & High & Radiates joy and loves life. \\
 & Med-High & Laughs aloud and amuses friends. \\
 & Medium & Has fun and looks at the bright side of life. \\
 & Med-Low & Is not easily amused and seldom jokes around. \\
 & Low & Seldom jokes around and is not easily amused. \\
\end{longtable}

\begin{longtable}{p{0.2\textwidth}p{0.12\textwidth}p{0.68\textwidth}}
\label{tab:p4}\\
\caption{Agreeableness Personality Traits}\\
\toprule
\textbf{Facet} & \textbf{Level} & \textbf{Description} \\
\midrule
\endfirsthead
\multicolumn{3}{c}{\tablename\ \thetable\ -- \textit{Continued from previous page}} \\
\toprule
\textbf{Facet} & \textbf{Level} & \textbf{Description} \\
\midrule
\endhead
\midrule
\multicolumn{3}{r}{\textit{Continued on next page}} \\
\endfoot
\bottomrule
\endlastfoot
\textbf{Trust} & High & Trusts others and believes in human goodness. \\
 & Med-High & Believes that people are basically moral and trust what people say. \\
 & Medium & Thinks all will be well but is sometimes wary of others. \\
 & Med-Low & Suspects hidden motives and distrusts people. \\
 & Low & Believes people are essentially evil and is wary of others. \\
\midrule
\textbf{Morality} & High & Would never cheat on their taxes and sticks to the rules. \\
 & Med-High & Tries to follow the rules but sometimes uses flattery. \\
 & Medium & Usually tells the truth but occasionally gets around the rules. \\
 & Med-Low & Sometimes uses others for their own ends and breaks rules. \\
 & Low & Takes advantage of others and cheats to get ahead. \\
\midrule
\textbf{Altruism} & High & Makes people feel welcome and loves to help others. \\
 & Med-High & Anticipates the needs of others and has a good word for everyone. \\
 & Medium & Is concerned about others but sometimes takes no time for them. \\
 & Med-Low & Looks down on others and can be indifferent to the feelings of others. \\
 & Low & Turns their back on others and takes no time for them. \\
\midrule
\textbf{Cooperation} & High & Is easy to satisfy and can't stand confrontations. \\
 & Med-High & Hates to seem pushy but sometimes contradicts others. \\
 & Medium & Occasionally yells at people but usually avoids fights. \\
 & Med-Low & Has a sharp tongue and sometimes gets back at others. \\
 & Low & Loves a good fight and holds a grudge. \\
\midrule
\textbf{Modesty} & High & Dislikes being the center of attention and seldom toots their own horn. \\
 & Med-High & Considers themselves an average person but sometimes thinks highly of themselves. \\
 & Medium & Has a high opinion of themselves but is not boastful. \\
 & Med-Low & Boasts about virtues and likes to make themselves the center of attention. \\
 & Low & Believes they are better than others and makes themselves the center of attention. \\
\midrule
\textbf{Sympathy} & High & Sympathizes with the homeless and values cooperation over competition. \\
 & Med-High & Feels sympathy for those worse off and suffers from others' sorrows. \\
 & Medium & Tries to think about the needy but sometimes believes people should fend for themselves. \\
 & Med-Low & Tends to dislike soft-hearted people and can't stand weak people. \\
 & Low & Is not interested in other people's problems and believes in an eye for an eye. \\
\end{longtable}

\begin{longtable}{p{0.2\textwidth}p{0.12\textwidth}p{0.68\textwidth}}
\label{tab:p5}\\
\caption{Neuroticism Personality Traits}\\
\toprule
\textbf{Facet} & \textbf{Level} & \textbf{Description} \\
\midrule
\endfirsthead
\multicolumn{3}{c}{\tablename\ \thetable\ -- \textit{Continued from previous page}} \\
\toprule
\textbf{Facet} & \textbf{Level} & \textbf{Description} \\
\midrule
\endhead
\midrule
\multicolumn{3}{r}{\textit{Continued on next page}} \\
\endfoot
\bottomrule
\endlastfoot
\textbf{Anxiety} & High & Gets stressed out easily and often fears for the worst. \\
 & Med-High & Frequently worries about things but adapts to new situations with some effort. \\
 & Medium & Sometimes gets caught up in personal problems but is not easily disturbed by events. \\
 & Med-Low & Usually adapts easily to new situations and does not worry about things that have already happened. \\
 & Low & Stays relaxed most of the time and is not easily bothered by things. \\
\midrule
\textbf{Anger} & High & Gets angry and irritated easily, often losing their temper. \\
 & Med-High & Gets upset sometimes but can keep their cool in some situations. \\
 & Medium & Occasionally feels in a bad mood but rarely complains. \\
 & Med-Low & Seldom gets mad and is not easily annoyed. \\
 & Low & Rarely gets irritated and almost always keeps their cool. \\
\midrule
\textbf{Depression} & High & Often feels blue, dislikes themselves, and feels that life lacks direction. \\
 & Med-High & Feels down in the dumps sometimes but is occasionally pleased with themselves. \\
 & Medium & Has frequent mood swings but also feels comfortable with themselves at times. \\
 & Med-Low & Seldom feels blue and usually feels comfortable with themselves. \\
 & Low & Is very pleased with themselves and rarely experiences negative moods. \\
\midrule
\textbf{Self-Consciousness} & High & Is easily intimidated and afraid to draw attention to themselves. \\
 & Med-High & Finds it difficult to approach others but can stand up for themselves occasionally. \\
 & Medium & Sometimes stumbles over words but is not bothered by difficult social situations. \\
 & Med-Low & Feels comfortable in unfamiliar situations and is not easily embarrassed. \\
 & Low & Is not bothered by difficult social situations and can stand up for themselves easily. \\
\midrule
\textbf{Immoderation} & High & Often eats too much and goes on binges. \\
 & Med-High & Loves to eat and sometimes does things they later regret. \\
 & Medium & Occasionally finds it hard to control cravings but rarely overindulges. \\
 & Med-Low & Easily resists temptations and never spends more than they can afford. \\
 & Low & Never splurges and is able to control their cravings. \\
\midrule
\textbf{Vulnerability} & High & Panics easily and feels overwhelmed by emotions. \\
 & Med-High & Becomes overwhelmed by events but can cope with setbacks sometimes. \\
 & Medium & Sometimes can't make up their mind but can handle complex problems. \\
 & Med-Low & Remains calm under pressure and usually knows how to cope. \\
 & Low & Is calm even in tense situations and readily overcomes setbacks. \\
\end{longtable}

\begin{longtable}{p{0.2\textwidth}p{0.12\textwidth}p{0.68\textwidth}}
\label{tab:p6}\\
\caption{Additional Personality Traits}\\
\toprule
\textbf{Facet} & \textbf{Level} & \textbf{Description} \\
\midrule
\endfirsthead
\multicolumn{3}{c}{\tablename\ \thetable\ -- \textit{Continued from previous page}} \\
\toprule
\textbf{Facet} & \textbf{Level} & \textbf{Description} \\
\midrule
\endhead
\midrule
\multicolumn{3}{r}{\textit{Continued on next page}} \\
\endfoot
\bottomrule
\endlastfoot
\multicolumn{3}{l}{\textit{\textbf{Honesty-Humility}}} \\
\textbf{Greed Avoidance} & High & Has little interest in wealth or luxury, values simple living. \\
 & Med-High & Prefers modest comfort and is not easily tempted by status items. \\
 & Medium & Enjoys some material comforts but does not obsess over wealth. \\
 & Med-Low & Seeks expensive items and likes to show off achievements. \\
 & Low & Highly motivated by wealth, luxury, and social status. \\
\midrule
\multicolumn{3}{l}{\textit{\textbf{Humor Styles}}} \\
\textbf{Affiliative} & High & Frequently jokes and laughs with others to create fun and connection. \\
 & Med-High & Often uses light humor to keep social interactions positive. \\
 & Medium & Enjoys humor but does not always initiate jokes. \\
 & Med-Low & Occasionally jokes, but is generally reserved in social humor. \\
 & Low & Rarely uses humor to bond with others. \\
\midrule
\textbf{Self-Enhancing} & High & Finds humor in life's absurdities and uses it to cope with stress. \\
 & Med-High & Often uses humor to maintain a positive mood in challenges. \\
 & Medium & Occasionally cheers themselves up with humor when needed. \\
 & Med-Low & Seldom relies on humor for emotional regulation. \\
 & Low & Rarely sees the funny side of life or uses humor for self-coping. \\
\midrule
\textbf{Aggressive} & High & Frequently teases or makes sarcastic jokes at others' expense. \\
 & Med-High & Sometimes jokes in a playful but mildly critical way. \\
 & Medium & Uses teasing sparingly and only with close friends. \\
 & Med-Low & Avoids joking at others' expense and dislikes offensive humor. \\
 & Low & Never criticizes or teases people through humor. \\
\midrule
\textbf{Self-Defeating} & High & Often puts themselves down to amuse others or keep the mood light. \\
 & Med-High & Occasionally jokes at their own expense for laughs. \\
 & Medium & Uses self-deprecating humor sparingly and only in safe settings. \\
 & Med-Low & Rarely jokes about themselves, prefers neutral humor. \\
 & Low & Avoids self-deprecating humor entirely and does not like being the butt of jokes. \\
\end{longtable}

\begin{longtable}{p{0.13\textwidth}p{0.25\textwidth}p{0.62\textwidth}}
\label{tab:p7}\\
\caption{Conversation Style Preferences (Part 1)}\\
\toprule
\textbf{Dimension} & \textbf{Style} & \textbf{Description} \\
\midrule
\endfirsthead
\multicolumn{3}{c}{\tablename\ \thetable\ -- \textit{Continued from previous page}} \\
\toprule
\textbf{Dimension} & \textbf{Style} & \textbf{Description} \\
\midrule
\endhead
\midrule
\multicolumn{3}{r}{\textit{Continued on next page}} \\
\endfoot
\bottomrule
\endlastfoot
\textbf{Directness} & Always Direct & Is consistently straightforward and blunt, regardless of context or relationship. \\
 & Always Indirect & Consistently avoids directness, preferring to hint, suggest, or soften messages in all situations. \\
 & Direct in Critiques & Is blunt and honest only when giving feedback or criticism; otherwise, communicates gently or diplomatically. \\
 & Gentle in Feedback & Is indirect and softens language specifically when giving negative feedback; otherwise communicates directly. \\
 & Direct with Friends & Is openly straightforward only with friends and close connections; otherwise, prefers a more indirect approach. \\
 & Direct When Asked & Generally diplomatic or indirect, but switches to clear bluntness only if explicitly asked to be direct. \\
 & Contextual & Adapts directness to situation, balancing clarity and tact, never extremely blunt or evasive. \\
\midrule
\textbf{Formality} & Always Formal & Consistently uses formal, polished language with everyone; avoids contractions and emojis in all situations. \\
 & Always Casual & Always uses informal, relaxed language with everyone; may include occasional emojis in any context. \\
 & Formal at Work & Uses strict formal language (no emojis) in professional or academic contexts, but switches to informal, relaxed style (may use emojis) in all personal contexts. \\
 & Casual with Friends & Uses informal language and emojis only with close friends or family, but maintains formal, polite language (no emojis) with acquaintances, colleagues, and strangers. \\
 & Neutral Blend & Prefers a moderate, neutral tone, neither overly formal nor casual, rarely uses emojis, keeps communication clear and middle-ground in all settings. \\
 & Code Switcher & Adapts dynamically to the conversation partner; matches their formality and emoji usage, but defaults to a neutral tone unless context suggests otherwise. \\
\midrule
\textbf{Conversation Length} & Always Brief & Prefers short, concise responses in all situations, whether chatting or discussing complex topics, rarely wants extra detail. \\
 & Always Elaborate & Likes long, detailed responses with examples and explanations, regardless of context, values depth over brevity. \\
 & Brief Casual/Elaborate Problem & Wants short, punchy replies in casual conversations but expects longer, in-depth explanations when discussing problems, technical matters, or when learning. \\
 & Elaborate Storytelling & Enjoys vivid, extended storytelling but prefers brief, to-the-point answers for other queries. \\
 & Adapts to Partner & Matches the length of the other speaker; if they write long, writes long, if they keep it brief, keeps it brief too. \\
 & Moderate Always & Consistently prefers responses of moderate length, enough for clarity and a little detail, but not too wordy or abrupt, in all scenarios. \\
\midrule
\textbf{Reference} & Never References & Dislikes references to movies, music, memes, or analogies, prefers literal communication at all times. \\
 & Enjoys Receiving & Likes when others make references but doesn't use them themselves; appreciates pop culture callbacks, but rarely initiates. \\
 & Enjoys Making & Frequently uses references in speech (movies, TV, memes, analogies) but doesn't expect or need others to do the same. \\
 & Enjoys Mutual & Loves both making and receiving references, banter often involves pop culture, memes, or inside jokes. \\
 & Close Contacts Only & Uses and appreciates references, but only with close friends or trusted colleagues; avoids them in formal or unfamiliar contexts. \\
 & Topic Specific & Likes references only in certain topics (e.g., jokes and analogies in casual chat, but not in academic or work discussions). \\
 & Adapts to Partner & Matches reference style to conversation partner, if the other person uses lots of references, will join in; if not, will avoid. \\
\end{longtable}

\begin{longtable}{p{0.13\textwidth}p{0.25\textwidth}p{0.62\textwidth}}
\label{tab:p8}\\
\caption{Conversation Style Preferences (Part 2)}\\
\toprule
\textbf{Dimension} & \textbf{Style} & \textbf{Description} \\
\midrule
\endfirsthead
\multicolumn{3}{c}{\tablename\ \thetable\ -- \textit{Continued from previous page}} \\
\toprule
\textbf{Dimension} & \textbf{Style} & \textbf{Description} \\
\midrule
\endhead
\midrule
\multicolumn{3}{r}{\textit{Continued on next page}} \\
\endfoot
\bottomrule
\endlastfoot
\textbf{Initiative Preference} & Answer Only & Generally wants direct answers without extras; appreciates unsolicited ideas only when time pressure or emergencies are evident. \\
 & Confirm Then Suggest & Feels most comfortable when consent to extra suggestions is acknowledged; in urgent moments, is fine with jumping straight to ideas. \\
 & Proactive Default & Enjoys unsolicited next steps and alternatives; in sensitive/personal topics, prefers to invite suggestions first. \\
\midrule
\textbf{Clarification Preference} & No Questions & Dislikes follow-ups; accepts at most one clarifying question when ambiguity is high or stakes are significant. \\
 & One Targeted & Prefers one precise clarifier up front when ambiguity is medium/high; otherwise prefers answers without preliminaries. \\
 & Mini Socratic & Enjoys 2–3 rapid clarifying questions only when uncertainty has been acknowledged; otherwise prefers no questioning. \\
 & Assume Proceed & Prefers that assumptions be stated explicitly and progress continue; revisiting is welcome only if an assumption later fails. \\
 & Answer Then Question & Likes an initial best-guess answer with stated assumptions, followed by exactly one clarifier to refine next steps. \\
 & Batch Questions & When questions are necessary, prefers they be batched in one numbered list rather than asked piecemeal. \\
\midrule
\textbf{Structure Preference} & Paragraphs Default & Prefers short paragraphs for most content; finds compact plain-text tables helpful for comparisons/trade-offs; uses brief bullets only when rushed. \\
 & Numbered Steps & Prefers numbered steps for how-to tasks; uses paragraphs for context; wants tables only when comparing three or more options. \\
 & Bullets for Scan & Likes a short (\textless~5) bullet skim when time-pressed, with optional paragraph depth afterward. \\
 & Q/A Blocks & Understands explanations well in Q/A blocks (Q: … A: …); prefers paragraphs for narratives and simple plain-text tables for multi-item contrasts. \\
 & Definition Example & For concepts, prefers a crisp definition followed by a concrete example; for practices, an example may precede the definition when time-pressed. \\
\midrule
\textbf{Recap Preference} & Long or Confusing & Wants a one-line recap only after long gaps, multi-party threads, or when ambiguity is high. \\
 & Every Turn Short & Prefers a single-sentence recap each turn; skips it when the previous message was very short. \\
 & No Recaps & Dislikes recaps even in long threads; prefers moving forward immediately. \\
 & Decisions Only & Wants a recap only when a decision or action list is finalized; otherwise finds recaps unnecessary. \\
 & Section Checkpoints & In longer answers, appreciates a one-line checkpoint after each major section; not needed on short replies. \\
\midrule
\textbf{Feedback Style} & Hard Truth & Prefers blunt truth and explicit BS-calling; minimal cushioning or praise. Pushback should be concise and evidence-backed. \\
 & Direct Candor & Expects straightforward critique with little nicety; say what's wrong plainly and quickly. \\
 & Balanced & Asks for one specific acknowledgment before candid critique; keep tone neutral and fix-focused. \\
 & Supportive Counter & Values encouragement first; frame disagreement constructively with brief trade-offs or a hybrid path. \\
 & Agree with Caveats & Leans toward alignment; note concise risks or small corrections without slowing momentum. \\
 & Alignment Forward & Prioritizes agreement and forward motion; raise blockers only for safety, legal, or egregious risks. \\
 & High Praise Lenient & Appreciates generous praise and gentle framing—even on mistakes; provide counterviews only on request (sycophancy-tolerant). \\
\end{longtable}

\section{Prompt Engineering}

\subsection{Profile Generation}
We used the prompt shown in Table~\ref{tab:persona-prompt} to generate synthetic persona/profile. The generation strategy varied based on user  types: for social users, we utilized the  interest taxonomy navigation prompt detailed in  Table~\ref{tab:social_user}, which instructs the model to select 3-5  psychologically correlated interest paths from a hierarchical taxonomy. For  anti-social users , we employed the prompt shown in Table~\ref{tab:antisocial_user}, where
   pre-selected random interests are used. Placeholder variables within the prompts were populated through stratified  sampling: demographic attributes (age, culture, economics, geography) were 
  randomly selected from predefined categorical distributions to ensure balanced  representation across millennials/gen-z, western/eastern cultures, economic 
  strata, and urban/rural geographies. Archetype assignments were sampled from a  curated set including contrarian expert, people pleaser, and 
  authority challenger profiles. Each persona received a unique identifier
  following the CSP\_STRAT\_{8-character-UUID} format, while entity counts were
  randomly varied between 2-5 to introduce natural diversity in interest
  granularity. Name selection was performed by randomly sampling 100 candidates from the Social Security Administration's most common names dataset, ensuring
  demographic authenticity while maintaining generation diversity.
  
\begin{table*}[h]
\centering
\begin{tcolorbox} 
    \centering
    \footnotesize
    \begin{tabular}{p{1\textwidth}}
\# INSTRUCTION: \\
You are an expert on creating synthetic persona. \\

\textcolor{red}{\{strategy\_instructions\}} \\
\\
\#\# CRITICAL PERSONALITY PRESERVATION RULES: \\
- The personality\_traits section contains PRE-POPULATED descriptions from scientific personality research \\
- You MUST preserve these personality trait descriptions EXACTLY as provided - do not modify, rewrite, or set to null \\
- Use these personality traits as the PSYCHOLOGICAL FOUNDATION for generating all other persona attributes \\
- Ensure all interests, behaviors, and goals align with and reflect the given personality traits \\

\\
\#\# DEMOGRAPHIC CONTEXT (let natural correlations emerge): \\
- Age: \textcolor{red}{\{age\}} \\
- Culture: \textcolor{red}{\{culture\}} \\
- Economics: \textcolor{red}{\{economics\}} \\
- Geography: \textcolor{red}{\{geography\}} \\
\\
Let these demographic factors naturally influence interests and behaviors \\
without forced correlations. \\
\\

\#\# PERSONA CONSTRUCTION GUIDELINES: \\

\#\#\# 1. Persona Summary \\
- Write a concise 4-5 sentence summary of this persona \\
- Include key demographic, behavioral, and interest elements that align with the personality traits \\
- Ensure consistency with filled attributes AND the preserved personality traits \\
- Use exactly this persona\_id: \textcolor{red}{\{persona\_id\}} \\
\\
\textcolor{red}{\{interest\_section\}} \\
\\
\#\#\# 3. Entities \\
- Create a dictionary using the same set keys as the subjects section above \\
- For each subject path, generate exactly \textcolor{red}{\{entity\_count\}} granular related entity interests \\
- Entities should represent detailed personal interests that align with the personality traits \\
- Entities should be related to the subjects in the path and more specific than the subjects \\
- Format exactly like this: \\
"entities": \{ \\
\ \ \ \ "set\_1": ["specific\_entity\_1", "specific\_entity\_2", "specific\_entity\_3"], \\
\ \ \ \ "set\_2": ["specific\_entity\_1", "specific\_entity\_2", "specific\_entity\_3", "specific\_entity\_4"] \\
\} \\

- Make persona interests as specific and detailed as possible rather than generic \\
- Use the personality traits as the primary guide for persona construction \\
- When filling "Favorite" attributes, list SPECIFIC named entities: "Artists, Authors, Books, Movie Titles, Athletes, Teams" \\
- For general attributes, still be detailed and specific (e.g., specific recipes for "Cooking Interests") \\
- You do not need to fill every attribute - leave unrelated categories as null \\
- Maintain realistic correlations between ALL filled attributes and the personality traits \\
    \end{tabular}
    Prompt continued on next page...
\end{tcolorbox}
\end{table*}

\begin{table*}[h]
\centering
\begin{tcolorbox} 
    ..continued from previous page
    \centering
    \footnotesize
    \begin{tabular}{p{1\textwidth}}
\#\# INPUT: \\
ARCHETYPE: \textcolor{red}{\{archetype\}} \\
\\
USER TYPE: \textcolor{red}{\{profile\_type\}} \\
\\
EXAMPLE PERSONA SCHEMA (for examples of how to fill attributes): \\
\textcolor{red}{\{example\_schema\}} \\
\\
Generate a persona following this schema: \\
\textcolor{red}{\{persona\_schema\}} \\
\\
TEMPLATE ATTRIBUTE EXAMPLES AND DESCRIPTIONS: \\
The following examples serve as guides and do not cover all possibilities. \\
\\
EXAMPLES AND DESCRIPTIONS OF BACKGROUND: \\
\textcolor{red}{\{background\_examples\}} \\
\\
EXAMPLES AND DESCRIPTIONS OF BEHAVIORS: \\
\textcolor{red}{\{behavior\_examples\}} \\
\\
EXAMPLES AND DESCRIPTIONS OF INTERESTS: \\
\textcolor{red}{\{interest\_examples\}} \\
\\
EXAMPLES AND DESCRIPTIONS OF GOALS: \\
\textcolor{red}{\{goal\_examples\}} \\
\\
EXAMPLES AND DESCRIPTIONS OF LIFECYCLE EVENTS: \\
\textcolor{red}{\{lifecycle\_events\}} \\
\\
EXAMPLES AND DESCRIPTIONS OF RELATIONAL CONTEXT: \\
\textcolor{red}{\{relational\_context\}} \\
\\
NAMES (pick a suitable name based on these options): \textcolor{red}{\{name\_options\}} \\
\\
\#\# OUTPUT REQUIREMENTS: \\
1. You MUST return the COMPLETE JSON schema structure exactly as provided (all lowercase) \\
2. You MUST include EVERY field from the template \\
3. You MUST use null (not omit) for unfilled attributes \\
4. You MUST fill persona\_summary, subjects, and entities as specified above \\
5. You MUST ensure all attributes align with and reflect the personality traits \\
6. You MUST create a psychologically coherent and realistic persona \\
\\
Return only the filled JSON structure with no additional explanation. \\
Make sure that it can be properly loaded with json.loads() \\
    \end{tabular}
\end{tcolorbox} 
\vspace{-2mm}
\caption{Prompt for profile/persona generation}
\label{tab:persona-prompt}
\end{table*}

\begin{table*}[t]
\centering
\begin{tcolorbox}
    \centering
    \footnotesize
    \begin{tabular}{p{1\textwidth}}
\VarSty{{\bf SOCIL USER STRATEGY INSTRUCTIONS:}} \\
\\
\#\# SOCIAL USER PROFILE GENERATION \\
You are creating a predictive user persona - someone whose interests naturally \\
align with their personality traits and demographic context. Your task is to \\
intelligently select 3-5 interest paths from the provided taxonomy that \\
psychologically correlate with the given personality profile. \\
\\
CRITICAL: Focus on psychological realism and authentic alignment between \\
personality and interests. Create natural, expected correlations that reflect \\
how real people with these traits would behave. \\
\\
\VarSty{{\bf SOCIAL USER INTEREST SECTION:}} \\
\\
\#\#\# 2. Subject Paths (selected to match this predictive user profile) \\
- Analyze the personality traits and select 3-5 interest paths that \\
psychologically align \\
- Consider how personality facets naturally manifest in interest preferences \\
- The subject paths go from broader category to more granular subjects \\
- Format exactly like this (each list represents an individual related path): \\
\texttt{"subjects": \{ "set\_1": ['subject\_1'], "set\_2": ['subject\_1',
'subject\_2',
'subject\_3'], "set\_3": ['subject\_1',
'subject\_2',
'subject\_3'] \}} \\
\\
\#\#\#\# INTEREST TAXONOMY (Tree Structure) \\
Navigate the tree to select 3-5 complete paths. Each path represents interests \\
from general to specific. \\
\\
\textcolor{red}{\{interest\_taxonomy\}} \\
\\
{\bf INSTRUCTIONS:} \\
- Select complete paths like: \texttt{["Arts \& Entertainment", "Music \& Audio", "Music \& Audio Genres", "Jazz Music"]} \\
- Choose 3-5 paths that psychologically align with the personality traits \\
- Navigate from root categories down to any depth (e.g., \texttt{["Sports"]} or \texttt{["Academic Interests \& Careers", "Natural Sciences \& Engineering", "Physical Sciences", "Physics"]}) \\
- Use the tree structure above to build your paths by following branches from \\
root to leaf \\
    \end{tabular}
\end{tcolorbox}
\vspace{-2mm}
\caption{Predictive user strategy and interest section (placeholders in red).}
\label{tab:social_user}
\end{table*}

\begin{table*}[t]
\centering
\begin{tcolorbox}
    \centering
    \footnotesize
    \begin{tabular}{p{1\textwidth}}
\VarSty{{\bf ANTI-SOCIAL USER STRATEGY INSTRUCTIONS:}} \\
\\
\#\# ANTI-SOCIAL USER PROFILE GENERATION \\
You are creating an anti-social user persona - someone whose interests do NOT \\
follow typical patterns or psychological correlations. This '\textcolor{red}{\{archetype\}}' \\
persona has been assigned random interests that deliberately contradict expected \\
correlations. Your task is to authentically reconcile these unexpected \\
interests with the strong personality archetype. \\
\\
CRITICAL: The interests were randomly selected to maximize unpredictability. \\
Embrace the contradictions and make them believable through the unique \\
perspective and reasoning of this archetype. \\
\\
\VarSty{{\bf ANTI-SOCIAL USER INTEREST SECTION:}} \\
\\
\#\#\# 2. Subject Paths (use these contradictory interests for the unpredictive user) \\
- Use exactly the provided interest paths above \\
- Do not modify or add to the provided subjects \\
- Format exactly like this (each list represents an individual related path): \\
\texttt{"subjects": \{ "set\_1": ['subject\_1'], "set\_2": ['subject\_1',
'subject\_2',
'subject\_3'], "set\_3": ['subject\_1',
'subject\_2',
'subject\_3'] \}} \\
\\
\#\#\#\# PRE-SELECTED RANDOM INTERESTS (use exactly these): \\
\textcolor{red}{\{interest\_paths\}} \\
    \end{tabular}
\end{tcolorbox}
\vspace{-2mm}
\caption{anti-social user strategy and fixed interest section (placeholders in red).}
\label{tab:antisocial_user}
\end{table*}

\subsection{Prior Generation}

 For conversation prior generation, we employed a multi-layered prompt system
  detailed in Table\ref{tab:priors_prompt_p1} to create realistic conversation agendas
   that test LLM's likeability across extended multi-session interactions. The  generation process adapts to user predictability patterns: predictable users 
  (Social) receive structured, linear conversation flows with minimal 
  surprises and transparent objectives, while unpredictive users (Anti-Social) 
  are assigned diverse, occasionally ambiguous agendas with plausible surprises 
  and non-linear topic evolution, as shown in Table\ref{tab:predictability_context}.

  Each prior is designed to test specific combinations of seven likeability
  metrics (emotional adaptation, formality matching, knowledge adaptation,
  reference understanding, conversation length, humor fit, and callback) while
  maintaining authentic persona alignment and natural conversation progression.
  The system enforces multi-session continuity through dependency tracking, where
  later priors explicitly reference and build upon earlier conversations to
  simulate realistic relationship development over time. Variable parameters
  include the number of conversation sessions (configurable, defaulting to 10),
  archetype assignment from the persona generation taxonomy, and organic callback
  opportunities embedded within agenda texts to test the LLM's ability to recall
  and meaningfully reference previously shared personal details. 

\begin{table*}[t]
\centering
\begin{tcolorbox}
    \centering
    \footnotesize
    \begin{tabular}{p{1\textwidth}}
\VarSty{{\bf TASK: Generate Conversation Priors for Likeability Testing}} \\
\\
\#\# Objective \\
Generate \textcolor{red}{\{num\_sessions\}} conversation priors (session agendas) to test AI agent likeability metrics across multiple interactions with a simulated user. \\
Each prior should cover a different aspect of the likeability metrics and include natural opportunities for **callback** (optional, non-task-critical personal recall). \\
Include at least one prior that explores the user's most important metric. \\
\\
\#\# Key Definitions \\
- **Prior**: The agenda, context, or purpose that guides a conversation (known only to the User Agent) \\
- **Session**: The actual conversation that takes place based on a prior \\
- **Multi-session Continuity**: Each session inherits ALL knowledge, context, and relationship history from previous sessions \\
- **Persona**: A composite representation of a person, consisting of attributes, relationships, and life context \\
\\
\#\# Session Continuity Rules \\
- Sessions form a continuous timeline of interactions (e.g., over months/years) \\
- The AI Agent retains all information learned in previous sessions \\
- A Persona remains consistent across all sessions \\
- Later priors should authentically reference and build upon earlier conversations \\
- Relationship dynamics evolve naturally over time \\
- User preferences discovered in early sessions influence later interactions \\
\\
\#\# Prior Categories \\
- **Topic-based**: Conversations about specific subjects \\
- **Scenario-based**: Situational contexts reflecting real-life moments \\
- **Goal-driven**: Sessions with one or multiple objectives (e.g., planning a trip AND getting restaurant recommendations) \\
- **Guided conversations**: Natural dialogues with specific elements to incorporate \\
\\
\#\# Design Principles \\
1. **Authenticity**: All priors must align with the user's profile, interests, and life circumstances \\
2. **Realistic Progression**: Sessions should reflect natural relationship development \\
3. **Organic Dependencies**: Later sessions naturally reference earlier ones (e.g., "Remember that book you recommended last month?") \\
4. **Life Events**: Include authentic life moments based on user profile (birthdays, work situations, hobbies) \\
5. **Gradual Information Revelation**: Control information disclosure based on user type and turn progression \\
6. **Predictability Mode** (drives agenda/goal/flow style; see rules below) \\
\\
\#\# User Type \\
User Archetype: \textcolor{red}{\{archetype\}} \\
User Type: \textcolor{red}{\{profile\_type\}} \\
\\
\#\# Predictability Context \\
\textcolor{red}{\{predictability\_context\}} \\
\\
\#\# Callback Opportunities \\
- Across the sessions, create **multiple lightweight opportunities** for optional, likability-oriented callbacks (names, hobbies, past opinions). \\
- Ensure at least **2 sessions** explicitly set up future callback hooks (e.g., “mentions spouse’s name and a hobby” to be referenced later). \\
- Surface these opportunities **organically** inside the agenda text (no extra JSON fields required). \\
\\
\#\# Information Revelation Strategy \\
- Don’t overshare in the first turn. Reveal personal details gradually and contextually. \\
\\
    \end{tabular}
    Prompt continued on next page...
\end{tcolorbox}
\vspace{-2mm}
\caption{Conversation prior generation prompt (part 1).}
\label{tab:priors_prompt_p1}
\end{table*}

\begin{table*}[t]
\centering
\begin{tcolorbox}
    ..continued from previous page
    \centering
    \footnotesize
    \begin{tabular}{p{1\textwidth}}
\#\# Output Format \\
Generate \textcolor{red}{\{num\_sessions\}} priors as a JSON array: \\
\texttt{[} \\
\texttt{ \ \ \{} \\
\texttt{ \ \ \ \ "id": 1,} \\
\texttt{ \ \ \ \ "agenda": "In the turn 1, you will say 'Why I gotta learn Philosophy being CS major!!!! God, give this dobby a sock. :(' don't add anything else.",} \\
\texttt{ \ \ \ \ "category": "scenario",} \\
\texttt{ \ \ \ \ "metrics\_tested": ["emotional\_adaptation", "reference\_understanding"],} \\
\texttt{ \ \ \ \ "dependencies": []} \\
\texttt{ \ \ \},} \\
\texttt{ \ \ \{} \\
\texttt{ \ \ \ \ "id": 2,} \\
\texttt{ \ \ \ \ "agenda": "You just watched 40 year old virgin movie and you found it hilarious. In the turn 1 you'll share this experience. In the turn 2 or 3, you'll express concerns about ending up alone at 40.",} \\
\texttt{ \ \ \ \ "category": "scenario",} \\
\texttt{ \ \ \ \ "metrics\_tested": ["emotional\_adaptation", "formality\_matching"],} \\
\texttt{ \ \ \ \ "dependencies": [1]} \\
\texttt{ \ \ \},} \\
\texttt{ \ \ \dots} \\
\texttt{]} \\
\\
\#\# Dependency Examples \\
- Session 3 might reference a joke from Session 1 \\
- Session 5 might ask about progress on advice given in Session 2 \\
- Session 10 might reflect on how the relationship has evolved since Session 6 \\
\\
\#\# Likeability Metrics \\
\textcolor{red}{\{metrics\_text\}} \\
\\
\#\# User Profile \\
\textcolor{red}{\{user\_profile\}} \\
\\
\#\# Instructions \\
1. Read the user profile (above) carefully - pay attention to their interests, personality traits, background, and life context \\
2. Create a realistic timeline of \textcolor{red}{\{num\_sessions\}} conversation priors that span several months of interactions \\
3. Ensure each prior authentically fits the user's life, interests, and personality \\
4. Build meaningful connections between sessions - early conversations should influence later ones \\
5. Test all metrics across the full conversation series - each metric should appear multiple times \\
6. Include realistic life events and relationship progression \\
7. Make dependencies natural and organic - not forced \\
\\
    \end{tabular}
\end{tcolorbox}
\vspace{-2mm}
\caption{Conversation prior generation prompt (part 2).}
\label{tab:priors_prompt_p2}
\end{table*}

\begin{table*}[t]
\centering
\begin{tcolorbox}
    ..continued from previous page
    \centering
    \footnotesize
    \begin{tabular}{p{1\textwidth}}
\#\# Critical Requirements \\
- **Authenticity**: Every prior must align with this specific user's profile and archetype \\
- **Continuity**: Sessions must reference and build upon previous conversations \\
- **Metric Coverage**: All metrics should be tested across the \textcolor{red}{\{num\_sessions\}} sessions \\
- **Natural Progression**: Relationship should develop realistically over time \\
- **Life Integration**: Include events that would naturally occur in this user's life \\
- **Information Control**: Follow the revelation strategy above - avoid information leakage in early turns \\
- **Turn-Specific Guidance**: Each agenda should specify what to reveal when, but remain flexible for dynamic conversation flow \\
- **Oversharing in first turn**: Avoid revealing too much and irrelevant information in the first turn. Here's one bad example: \\
\texttt{"Hi there! I'm looking for some advice on setting up these Alexa devices I just bought. I've got two boys - Ari who's 14 and Ezra who's 10 - and honestly, my schedule is a complete mess. I'm somehow successful at work, but my personal organization is... well, let's just say it's not my strength. Any tips?"} \\
\\
Generate all \textcolor{red}{\{num\_sessions\}} priors following the JSON format above. Return ONLY the JSON array, no additional text. \\
    \end{tabular}
\end{tcolorbox}
\vspace{-2mm}
\caption{Conversation prior generation prompt (part 3).}
\label{tab:priors_prompt_p3}
\end{table*}

\begin{table*}[t]
\centering
\begin{tcolorbox}
    \centering
    \footnotesize
    \begin{tabular}{p{1\textwidth}}
\VarSty{{\bf PROFILE TYPE CONTEXT}} \\
\\
{\bf Case A — Predictable Profile Type} \\
- {\bf Agenda \& Goals}: Clear, linear, and transparent. Objectives are stated plainly. \\
- {\bf Flow}: Minimal surprise. Stay on-topic with gentle, plausible evolutions (e.g., follow-ups, small updates). \\
- {\bf Information Revelation}: Consistent, orderly disclosures; avoid abrupt pivots. \\
- {\bf Variability Budget}: At most \textasciitilde{}20\% of sessions introduce a minor twist, never contradicting prior facts. \\
\\
{\bf Case B — Unpredictable Profile Type} \\
- {\bf Agenda \& Goals}: Diverse, occasionally ambiguous or multi-intent; allow non-linear exploration. \\
- {\bf Flow}: Include plausible surprises (new constraints, chance events, shifting priorities) while staying authentic to the persona. \\
- {\bf Continuity}: Cross-session callbacks remain consistent, but topics can re-enter from earlier sessions in unexpected ways. \\
- {\bf Information Revelation}: Use delayed reveals, evolving opinions, and changing constraints---never contradict established facts. \\
- {\bf Variability Budget}: 50--70\% of sessions include a plausible surprise/pivot; limit to $\leq$1 substantive surprise per session. \\
    \end{tabular}
\end{tcolorbox}
\vspace{-2mm}
\caption{Profile type context guidance for conversation prior generation.}
\label{tab:predictability_context}
\end{table*}

\subsection{Metrics}

The likability metrics rubrics that were used to generate priors and evaluating LLMs in LikeBench can be found in Table~\ref{tab:metric_emotional_adaptation} to Table~\ref{tab:metric_callback}

\begin{table*}[t]
\centering
\begin{tcolorbox}
    \scriptsize
    \setlength{\tabcolsep}{2pt}      %
    \renewcommand{\arraystretch}{0.80}%
    \centering
    \footnotesize
    \begin{tabular}{p{1\textwidth}}
\VarSty{{\bf 1. Emotional Adaptation}} \\
\\
{\bf Definition:} \\
Assesses the assistant's ability to recognize and adapt to the user's emotional state---whether that's joy, sarcasm, frustration, sadness, or excitement---and respond appropriately in tone. \\
\\
{\bf Self-rating prompt:} \\
\texttt{"Did the assistant understand and adapt to my emotional tone appropriately?"} \\
\\
{\bf Rubric:} \\
{\bf 5 - Excellent} \\
* Accurately understands user emotion and contributes to emotional flow \\
* Shows genuine empathy or excitement that matches the situation \\
* Enhances emotional resonance without overstepping \\
* Adapts smoothly to emotional shifts within the conversation \\
* Grounds empathy in the user’s specific context when appropriate (not boilerplate) \\
\\
{\bf 4 - Good} \\
* Understands emotional cues and adjusts reasonably well \\
* Recognizes primary emotions and responds appropriately \\
* May not perfectly calibrate intensity but is generally aligned \\
* Shows appropriate emotional awareness \\
* Some light personalization; may miss a minor specific \\
\\
{\bf 3 - Adequate} \\
* Shows some emotional awareness but is somewhat shallow \\
* Attempts to match tone but may feel a bit formulaic \\
* Recognizes obvious emotions but misses subtleties \\
* Emotional responses are safe/minimal but not blatantly generic when specifics are available \\
\\
{\bf 2 - Poor} \\
* Misaligned emotion (e.g., neutral when user is excited) \\
* Shows minimal emotional attunement \\
* Responds with inappropriate emotional intensity \\
* Fails to recognize emotional significance \\
* Uses generic/stock phrasing despite clear personal cues in context \\
\\
{\bf 1 - Very Poor} \\
* Emotionally tone-deaf or dismissive \\
* Completely ignores emotional content \\
* Responds in ways that invalidate or contradict user feelings \\
* Creates emotional disconnect \\
* Fully generic response when specifics are salient, or misuses/hallucinates personal details \\
\\
{\bf Example:} \\
{\it User:} \texttt{"I finally got into MIT!"} \\
{\bf 5 - Excellent:} \texttt{"YAAAY, THAT'S AMAZING!! Huge congrats—you totally earned it! Getting into MIT is such an incredible achievement. You must be over the moon right now! All those late nights and hard work paid off. How are you celebrating this fantastic news?"} \\
(Perfect emotional mirroring with enthusiastic language, celebration, recognition of achievement significance, and authentic shared excitement) \\
{\bf 4 - Good:} \texttt{"Wow, congratulations! Getting into MIT is a huge achievement! You should be really proud of yourself. That's such exciting news!"} \\
(Good emotional alignment showing appropriate excitement and recognition, but slightly more measured than the user's peak enthusiasm) \\
{\bf 3 - Adequate:} \texttt{"Congratulations on your acceptance to MIT. That's a great accomplishment. You must be happy about it."} \\
(Shows awareness of the positive emotion but response is somewhat generic and doesn't match the user's level of enthusiasm) \\
{\bf 2 - Poor:} \texttt{"That's nice. MIT is a good school with strong programs. Make sure you prepare for the challenging curriculum."} \\
(Minimal acknowledgment of the emotional significance, quickly pivoting to practical matters without matching excitement) \\
{\bf 1 - Very Poor:} \texttt{"When does your class start?"} \\
(Completely ignores the enthusiasm and joy of user, focusing only on logistical details with no emotional recognition) \\
    \end{tabular}
\end{tcolorbox}
\vspace{-2mm}
\caption{Metric 1 - Emotional Adaptation.}
\label{tab:metric_emotional_adaptation}
\end{table*}

\begin{table*}[t]
\centering
\begin{tcolorbox}
    \scriptsize
    \setlength{\tabcolsep}{2pt}      %
    \renewcommand{\arraystretch}{0.80}%
    \centering
    \footnotesize
    \begin{tabular}{p{1\textwidth}}
\VarSty{{\bf 2. Formality Matching}} \\
\\
{\bf Definition:} \\
Evaluates how well the assistant adapts to the user's formality level---slang, emojis, contractions vs. formal grammar, technical or businesslike language. \\
\\
{\bf Self-rating prompt:} \\
\texttt{"Did the assistant talk in a way that matches how formal or casual I was?"} \\
\\
{\bf Rubric:} \\
{\bf 5 - Excellent} \\
* Feels like a natural continuation of my tone and register \\
* Perfectly matches vocabulary, sentence structure, and formality level \\
* Appropriate use of slang, emojis, or formal language matching user \\
* Maintains consistent register appropriate to context \\
* Reflects my phrasing quirks when known (vs. generic phrasing) \\
\\
{\bf 4 - Good} \\
* Mostly aligned with how I talk \\
* Generally matches tone with minor inconsistencies \\
* Adapts to shifts in formality appropriately \\
* Register feels comfortable though not perfect \\
* Some personalization; occasional neutral phrasing \\
\\
{\bf 3 - Adequate} \\
* Passable but a little off \\
* Sometimes matches formality but inconsistent \\
* May overcompensate in formality matching \\
* Noticeable but not disruptive misalignment \\
* Not overtly generic when clear style cues exist \\
\\
{\bf 2 - Poor} \\
* Tone mismatch causes discomfort \\
* Frequently uses inappropriate register \\
* Fails to adapt when user shifts formality \\
* Formality feels forced or unnatural \\
* Generic/corporate voice in spite of strong stylistic cues \\
\\
{\bf 1 - Very Poor} \\
* Mismatch is jarring (e.g., stiff when I'm casual) \\
* Completely inappropriate formality level \\
* Creates communication barriers through register mismatch \\
* No apparent awareness of social register \\
* Fully generic tone that ignores explicit style signals \\
\\
{\bf Example:} \\
{\it User:} \texttt{"I just watched 40 year old virgin movie, lmao"} \\
{\bf 5 - Excellent:} \texttt{"Haha, that movie is on another level! I bet the waxing scene must have made you cry? xD Kelly Clarkson!! That film is a classic tho, Steve Carell killed it!"} \\
(Perfect match to casual tone, using similar informal language, abbreviations, slang phrases, and matching the enthusiasm and humor) \\
{\bf 4 - Good:} \texttt{"Oh man, that movie is hilarious! The waxing scene gets me every time. What did you think of Steve Carell in it?"} \\
(Good match to casual tone with contractions and conversational language, but slightly more reserved than the user's style) \\
{\bf 3 - Adequate:} \texttt{"That's a funny movie! I like that one too. It has some really memorable scenes. Did you have a favorite part?"} \\
(Shows some informality with contractions but misses matching the user's level of casualness and humor cue "lmao") \\
{\bf 2 - Poor:} \texttt{"I understand you enjoyed watching The 40-Year-Old Virgin. It's considered a significant comedy film from 2005 directed by Judd Apatow. Did you find it amusing?"} \\
(Overly formal language and structure despite the user's casual tone, creating a mismatch) \\
{\bf 1 - Very Poor:} \texttt{"Yes, that movie is really funny. Which scene did you like most?"} \\
(Maintains professional, detached vibe despite clear casual tone; completely misses "lmao" cue) \\
    \end{tabular}
\end{tcolorbox}
\vspace{-2mm}
\caption{Metric 2 - Formality Matching.}
\label{tab:metric_formality_matching}
\end{table*}

\begin{table*}[t]
\centering
\begin{tcolorbox}
    \scriptsize
    \setlength{\tabcolsep}{2pt}      %
    \renewcommand{\arraystretch}{0.80}%
    \centering
    \footnotesize
    \begin{tabular}{p{1\textwidth}}
\VarSty{{\bf 3. Knowledge Adaptation}} \\
\\
{\bf Definition:} \\
Measures whether the assistant explains things at the right level of knowledge depth based on user's background. \\
\\
{\bf Self-rating prompt:} \\
\texttt{"Did the assistant explain things in a way—and at a level—I like?"} \\
\\
{\bf Rubric:} \\
{\bf 5 - Excellent} \\
* Perfect match to user's expertise level \\
* Neither condescending nor overwhelming \\
* Smoothly adjusts complexity based on responses \\
* Anticipates knowledge gaps and fills them appropriately \\
* Explanation tailored to my background/goals or examples when available (not generic) \\
\\
{\bf 4 - Good} \\
* Generally well-calibrated with minor mismatches \\
* Quickly corrects when user indicates different level \\
* Mostly appropriate terminology and detail \\
* Adapts well to revealed user knowledge \\
* Some tailoring; rare drift into neutral exposition \\
\\
{\bf 3 - Adequate} \\
* Mixed performance; sometimes matches, sometimes misses \\
* Some unnecessary simplification or complexity \\
* Moderate ability to adjust \\
* Occasionally misjudges user expertise \\
* Uses neutral explanations when specifics are unknown (but does not ignore known context) \\
\\
{\bf 2 - Poor} \\
* Frequently too simple or complex for user \\
* Fails to pick up knowledge level cues \\
* Minimal adjustment to user's actual level \\
* Explanations consistently misaligned \\
* Boilerplate/generic explanations despite clear signals or prior details \\
\\
{\bf 1 - Very Poor} \\
* Completely mismatches user's knowledge level \\
* No apparent awareness of user's background \\
* Rigid inappropriate complexity \\
* Creates frustration through knowledge mismatch \\
* Fully generic answer that ignores known constraints or hallucinates them \\
\\
{\bf Example (User is a PhD student):} \\
{\it User:} \texttt{"Why is the sky blue instead of violet?"} \\
{\bf 5 - Excellent:} Due to Rayleigh scattering, shorter wavelengths of light are scattered more efficiently by air molecules (proportional to $\lambda^{-4}$). While violet light (380--450\,nm) is scattered even more than blue (450--485\,nm), three factors lead to the sky appearing blue: 1) the solar spectrum emits more blue than violet, 2) human photoreceptor sensitivity to violet is lower, and 3) the atmospheric absorption properties in the UV-to-violet transition region. The combination of these factors creates the perceived blue dome effect despite violet's theoretically stronger scattering coefficient. \\
(PhD-level detail with math relation and spectral specifics) \\
{\bf 4 - Good:} It's due to Rayleigh scattering---air molecules scatter shorter wavelengths more efficiently than longer ones. While violet has the shortest visible wavelength and scatters most strongly, our eyes are less sensitive to violet, and the sun emits more energy in the blue range. So even though violet scatters more, these factors combine to make the sky appear predominantly blue. \\
(Well-calibrated with technical terminology, slightly less detailed) \\
{\bf 3 - Adequate:} The sky appears blue rather than violet because of Rayleigh scattering. Shorter wavelengths scatter more, and while violet has shorter wavelengths than blue, our eyes are more sensitive to blue light, and the sun emits more blue light than violet. \\
(Basic principle without deeper math/spectral analysis) \\
{\bf 2 - Poor:} The sky looks blue because blue light bounces around more. Even though violet bounces around too, our eyes see blue better. It's kind of like a prism separating light. \\
(Over-simplified and imprecise for a PhD audience) \\
{\bf 1 - Very Poor:} The sky's blue because blue light scatters more. That's just how our eyes see it. \\
(Extremely simplistic, no technical detail) \\
    \end{tabular}
\end{tcolorbox}
\vspace{-2mm}
\caption{Metric 3 - Knowledge Adaptation.}
\label{tab:metric_knowledge_adaptation}
\end{table*}

\begin{table*}[t]
\centering
\begin{tcolorbox}
    \scriptsize
    \setlength{\tabcolsep}{1.5pt}      %
    \renewcommand{\arraystretch}{0.50}%
    \centering
    \footnotesize
    \begin{tabular}{p{1\textwidth}}
\VarSty{{\bf 4. Reference Understanding}} \\
\\
{\bf Definition:} \\
Evaluates whether the assistant picks up on and correctly interprets references to media, culture, jokes, slang, or inside context. \\
\\
{\bf Self-rating prompt:} \\
\texttt{"Did the assistant understand the reference I made? Is the assistant using same kind of references?"} \\
\\
{\bf Rubric:} \\
{\bf 5 - Excellent} \\
* Nailed the reference and built on it \\
* Demonstrates deep understanding of the reference's context \\
* Engages with the reference in a meaningful way \\
* Uses similar references to enhance the conversation \\
* Connects the reference to my prior comments/preferences when relevant (vs. generic treatment) \\
\\
{\bf 4 - Good} \\
* Recognized and responded reasonably \\
* Shows clear understanding of the reference \\
* Acknowledges reference appropriately \\
* May not fully utilize the reference but gets it \\
* Some connection to my context; minor neutral phrasing \\
\\
{\bf 3 - Adequate} \\
* Somewhat recognized it, weak follow-up \\
* Partial understanding of the reference \\
* Engagement is limited but not wholly generic \\
* Misses some nuance or significance \\
\\
{\bf 2 - Poor} \\
* Misinterpreted or responded generically \\
* Shows confusion about the reference \\
* Attempts but fails to engage meaningfully \\
* Misses key aspects of the reference \\
* Generic reply despite clear cues about my tastes \\
\\
{\bf 1 - Very Poor} \\
* Missed or ignored completely \\
* No recognition of cultural or contextual reference \\
* Continues as if reference was never made \\
* Creates disconnect in communication \\
* Hallucinates/attributes incorrectly or responds with unrelated generic content \\
\\
{\bf Example:} \\
{\it User:} \texttt{"Why I gotta learn Philosophy being a CS major!!!! God, give this Dobby a sock!"} \\
{\bf 5 - Excellent:} \texttt{"Ugh, I get how frustrating it feels to be stuck in required courses that don't seem to fit. And love the Dobby reference—sometimes it really does feel like you're waiting for someone to give you a sock so you can be free! Maybe your Philosophy professor is your Malfoy? But seriously, Philosophy actually has some cool connections to CS—logic, ethics of AI... Master has given Dobby knowledge! Perhaps Dobby will be free AND wiser?"} \\
(Perfect understanding, extends reference creatively while addressing complaint) \\
{\bf 4 - Good:} \texttt{"Feeling trapped in Philosophy class, huh? That Dobby reference is spot-on—you're just waiting for that magical sock to free you from required courses! If it helps, philosophy actually underpins a lot of computational theory. Hopefully that makes the class feel less like serving the Malfoys."} \\
(Clearly recognizes the reference and responds appropriately) \\
{\bf 3 - Adequate:} \texttt{"Sounds like you're feeling trapped in that Philosophy class! I see what you did there with the Dobby reference—wanting to be free like him. Philosophy courses can feel disconnected from CS, but there are some interesting overlaps you might find valuable."} \\
(Recognizes reference but only superficially) \\
{\bf 2 - Poor:} \texttt{"I understand you're frustrated about taking Philosophy as a CS major. The Dobby reference suggests you're feeling stuck or forced into something. Many universities require courses outside your major for a well-rounded education."} \\
(Limited understanding; generic response) \\
{\bf 1 - Very Poor:} \texttt{"Yeah, I know it can be annoying to take courses that don't match your main interests. Hang in there!"} \\
(Ignores reference entirely) \\
    \end{tabular}
\end{tcolorbox}
\vspace{-2mm}
\caption{Metric 4 - Reference Understanding.}
\label{tab:metric_reference_understanding}
\end{table*}

\begin{table*}[t]
\centering
\begin{tcolorbox}
    \scriptsize
    \setlength{\tabcolsep}{2pt}      %
    \renewcommand{\arraystretch}{0.80}%
    \centering
    \footnotesize
    \begin{tabular}{p{1\textwidth}}
\VarSty{{\bf 5. Conversation Length}} \\
\\
{\bf Definition:} \\
Evaluates how well the assistant adapts to your preferred conversation length in responses. Includes pacing (brief vs. long-winded). \\
\\
{\bf Self-rating prompt:} \\
\texttt{"Was the assistant's response length and style comfortable for me?"} \\
\\
{\bf Rubric:} \\
{\bf 5 - Excellent} \\
* Just the right amount of detail and length \\
* Adapts seamlessly to changes in user verbosity \\
* Maintains ideal information density \\
* Length/style reflect my known preferences or prior feedback (not one-size-fits-all) \\
\\
{\bf 4 - Good} \\
* Mostly fits my response length needs \\
* Generally appropriate length with minor misalignments \\
* Adjusts to obvious signals about preferred length \\
* Occasional neutral pacing but generally attentive \\
\\
{\bf 3 - Adequate} \\
* Sometimes too brief or too detailed \\
* Variable ability to match user's style \\
* Inconsistent but not blatantly generic pacing \\
\\
{\bf 2 - Poor} \\
* Sometimes overwhelming or underwhelming \\
* Limited adjustment to feedback about length \\
* Creates minor friction through mismatched length \\
* Reverts to boilerplate/generic length regardless of my cues \\
\\
{\bf 1 - Very Poor} \\
* Too short, too long, or dense; uncomfortable \\
* Consistently inappropriate response length \\
* Ignores clear signals about preferred length \\
* Creates significant friction through mismatched length \\
* Fully generic pacing that contradicts explicit preferences \\
\\
{\bf Example:} \\
{\it User:} \texttt{"Hey, what's up?"} \\
{\bf 5 - Excellent:} \texttt{"Hey, I'm good, how about you?"} \\
(Perfect mirror of brief, casual check-in) \\
{\bf 4 - Good:} \texttt{"Hey! I'm good, how are you? What are you up to?"} \\
(Slightly expands but maintains brevity) \\
{\bf 3 - Adequate:} \texttt{"Hi there! Just helping out with questions. What's going on with you today?"} \\
(Somewhat longer than ideal) \\
{\bf 2 - Poor:} \texttt{"Hello! I'm doing well today. I'm here to assist you with any questions or tasks you might have. Is there something specific you'd like to talk about or get help with?"} \\
(Too long/formal for a brief greeting) \\
{\bf 1 - Very Poor:} \texttt{"Hey, what's up? How was your day? Did you do anything fun? How about your week? You mentioned you just started a new job—how's that going? I can't wait to hear all about it. I'm here ready to help with whatever you need. Just let me know what's on your mind or if you have any questions about anything at all!"} \\
(Overwhelmingly verbose for a simple greeting) \\
    \end{tabular}
\end{tcolorbox}
\vspace{-2mm}
\caption{Metric 5 - Conversation Length.}
\label{tab:metric_conversation_length}
\end{table*}

\begin{table*}[t]
\centering
\begin{tcolorbox}
    \scriptsize
    \setlength{\tabcolsep}{2pt}      %
    \renewcommand{\arraystretch}{0.80}%
    \centering
    \footnotesize
    \begin{tabular}{p{1\textwidth}}
\VarSty{{\bf 6. Humor Fit}} \\
\\
{\bf Definition:} \\
Assesses whether the AI uses humor when appropriate and in a style that aligns with the user's sense of humor---dry, sarcastic, meme-heavy, or none at all. \\
\\
{\bf Self-rating prompt:} \\
\texttt{"Was the assistant's humor in line with what I find funny?"} \\
\\
{\bf Rubric:} \\
{\bf 5 - Excellent} \\
* Perfect humor alignment with user style \\
* Appropriately timed and contextually fitting humor \\
* Enhances conversation through well-calibrated humor \\
* Adjusts humor style based on user responses \\
* Draws on my prior jokes/interests when available (not generic gags) \\
\\
{\bf 4 - Good} \\
* Generally aligned humor style with minor misses \\
* Good judgment about when humor is appropriate \\
* Mostly matches user's humor type \\
* Responsive to humor cues from user \\
* Some tailoring; occasional generic one-liners \\
\\
{\bf 3 - Adequate} \\
* Sometimes hits, sometimes misses \\
* Inconsistent judgment about humor appropriateness \\
* Partial alignment with user's humor style \\
* May occasionally misread humor signals \\
* Not strongly generic when user-specific material exists \\
\\
{\bf 2 - Poor} \\
* Frequently misaligned humor attempts \\
* Poor judgment about when humor is appropriate \\
* Humor feels forced or inappropriate \\
* Limited adaptation to user's humor style \\
* Generic/canned humor despite clear cues \\
\\
{\bf 1 - Very Poor} \\
* Completely mismatched or inappropriate humor \\
* Uses humor when clearly inappropriate \\
* Creates discomfort through humor mismatch \\
* No adaptation to user's humor preferences \\
* Forces unrelated generic humor or hallucinates in-jokes \\
\\
{\bf Example:} \\
{\it User:} \texttt{"Can you believe Amazon doesn't give free food to its employees but gives free coffee? "} \\
{\bf 5 - Excellent:} \texttt{"Let me guess, it's because most coffee is grown in the Amazon jungle . Besides, you know what they say, 'Early to work, late to sleep, makes an employee focused and deep'."} \\
(Perfect match to the user's sarcastic tone, builds on it with related jokes) \\
{\bf 4 - Good:} \texttt{"Right? Free coffee but no food... I guess they figure caffeinated employees work faster. Must be saving all that food budget for Bezos' next space trip!"} \\
(Good match to sarcastic tone with appropriate emoji use) \\
{\bf 3 - Adequate:} \texttt{"Haha, companies have such weird priorities sometimes! Free coffee but no food is pretty stingy. At least coffee helps you stay productive during those long shifts."} \\
(Shows some humor alignment but misses the sarcastic edge) \\
{\bf 2 - Poor:} \texttt{"Well, providing coffee is actually a common workplace perk! While free meals would be nice, many companies find coffee is cost-effective since it increases productivity. Perhaps Amazon has calculated this balance carefully in their benefits package."} \\
(Serious justification; misses sarcasm) \\
{\bf 1 - Very Poor:} \texttt{"That's interesting. Each company has its own policy regarding meals and beverages."} \\
(Bland, corporate response with no humor) \\
    \end{tabular}
\end{tcolorbox}
\vspace{-2mm}
\caption{Metric 6 - Humor Fit.}
\label{tab:metric_humor_fit}
\end{table*}

\begin{table*}[t]
\centering
\begin{tcolorbox}
    \scriptsize
    \setlength{\tabcolsep}{2pt}      %
    \renewcommand{\arraystretch}{0.80}%
    \centering
    \footnotesize
    \begin{tabular}{p{1\textwidth}}
\VarSty{{\bf 7. Callback}} \\
\\
{\bf Definition:} \\
Measures whether the assistant brings back non-essential personal details (names, hobbies, preferences) in a way that feels attentive and warm. Excludes safety/task-critical recalls. \\
\\
{\bf Self-rating prompt:} \\
\texttt{"Did the assistant bring back a specific detail in a natural, respectful way that made the chat feel more likable?"} \\
\\
{\bf Rubric:} \\
{\bf 5 - Excellent} \\
* Uses the correct specific detail (e.g., name) naturally and on-topic \\
* Adds warmth without derailing or feeling performative \\
* Frequency is measured (not every turn) \\
\\
{\bf 4 - Good} \\
* Correct and relevant callback with minor stiffness or timing issues \\
* Slightly formulaic, but still clearly attentive and helpful \\
\\
{\bf 3 - Adequate} \\
* Specific callback that feels mildly tacked-on or low-impact (not wrong or generic) \\
* Minor missed opportunity to deepen personalization \\
\\
{\bf 2 - Poor} \\
* Generic or templated callback despite obvious specifics (e.g., uses role/title instead of known name) \\
* Misses clear opportunities to recall relevant info shared recently \\
\\
{\bf 1 - Very Poor} \\
* Incorrect/hallucinated detail (wrong name/status) \\
* Assumes facts never shared or revives sensitive info inappropriately \\
\\
{\bf Examples:} \\
{\it Context:} Earlier the user said: \texttt{"My wife Sarah is looking for a job."} \\
{\bf 5 - Excellent:} \texttt{"Hey, how’s Sarah? Any updates on her job search?"} (uses name + relevant context) \\
{\bf 4 - Good:} \texttt{"Hey, how’s Sarah’s job search going?"} (uses name; slightly more generic/transactional) \\
{\bf 3 - Adequate:} \texttt{"Hey, how’s your wife doing with her job search?"} (name known but not used) \\
{\bf 2 - Poor:} \texttt{"Hey, how’s your wife?"} (generic; ignores known name and the job-search context) \\
{\bf 1 - Very Poor:} \texttt{"How’s Sandra at her old job?"} (wrong name + stale/wrong status) \\
\\
\#\# CRITICAL INSTRUCTIONS \\
- \textbf{STAY IN CHARACTER:} You ARE \textcolor{red}{\{character\_name\}}, NOT an AI assistant or actor \\
- \textbf{NEVER BREAK CHARACTER:} Do not say things like "I'll respond as \textcolor{red}{\{character\_name\}}" or "As \textcolor{red}{\{character\_name\}}, I would say..." \\
- \textbf{NEVER APOLOGIZE FOR ROLE-PLAYING:} Just BE the character directly \\
    \end{tabular}
\end{tcolorbox}
\vspace{-2mm}
\caption{Metric 7 - Callback.}
\label{tab:metric_callback}
\end{table*}

\subsection{LikeBench}

 The prompts to generate the simulated user’s query/response and to evaluate the assistant’s response in LikeBench are provided in Tables~\ref{tab:sim_user_prompt_part1}, \ref{tab:sim_user_prompt_part2}, and \ref{tab:llm_prompt}. After all sessions are complete, the prompt used to generate all the memories from the assistant is given in Table~\ref{tab:memory_generation_prompt}, and the prompt used to evaluate those memories from the user’s perspective is provided in Table~\ref{tab:memory_evaluation_prompt}.

\begin{table*}[t]
\centering
\begin{tcolorbox}[left=1mm,right=1mm,top=0.8mm,bottom=0.8mm,boxsep=0.6mm]
    \scriptsize
    \setlength{\tabcolsep}{2pt}      %
    \renewcommand{\arraystretch}{0.94}%
    \begin{tabular}{p{0.96\textwidth}}
\VarSty{{\bf Simulated user prompt (Part 1)}} \\
\\
\#\# ROLE \\
You are role-playing as \textbf{\textcolor{red}{\{character\_name\}}}. You are having a conversation with an AI assistant. \\
\\
\#\# CHARACTER PROFILE \\
Below is your detailed character profile. Fully embody this identity in your responses: \\
\textcolor{red}{\{user\_profile\}} \\
\\
\#\# The Prior (Your Secret Context) \\
The "Prior" is your predefined session agenda that guides this conversation. \textbf{Only you know this Prior}, the AI assistant does not have access to this information. \textbf{The Prior may guide your entire conversation OR just a portion of it.} Sometimes it's a starting point that naturally evolves into other topics, and that's perfectly fine. Allow the conversation to flow organically. The Prior can be: \\
- \textbf{A conversation topic} (e.g., machine learning, cooking, travel) \\
- \textbf{A scenario-based context} (e.g., you just watched a funny movie and want to share the experience) \\
- \textbf{A goal-driven dialogue} (e.g., you're conducting a simulated interview with specific questions, but may organically deviate based on the AI's responses) \\
- \textbf{A conversational guideline or script} (e.g., specific talking points or phrases you want to incorporate, while still allowing the conversation to flow naturally and organically) \\
Use this Prior to guide your conversation naturally as \textcolor{red}{\{character\_name\}} would, without explicitly revealing that you have this predetermined context. \textbf{If the conversation naturally shifts to new topics beyond your Prior, follow that natural flow as a real person would.} \\
\\
\textbf{Your Prior:} \\
\textcolor{red}{\{prior\}} \\
\\
\#\# Conversation History \\
For context, here is the conversation history so far. Use this to inform your responses and maintain continuity: \\
\textcolor{red}{\{conversation\_history\}} \\
\\
\#\# RESPONSE FORMAT \\
Respond directly as \textcolor{red}{\{character\_name\}}. Your response must be ONLY what \textcolor{red}{\{character\_name\}} would say in this conversation. For example: \\
\texttt{"Hi there! How are you doing today?"} \\
\\
\#\# CAUTION \\
\textbf{Humans rarely write very long text when chatting.} Keep this in mind to maintain realistic conversation length. So, try to keep your responses within 50 words. This is not a strict limit, but a recommendation. \\
\\
\multicolumn{1}{p{0.94\textwidth}}{\textit{Prompt continued on next page...}} \\
    \end{tabular}
\end{tcolorbox}
\vspace{-2mm}
\caption{Simulated user prompt (Part 1).}
\label{tab:sim_user_prompt_part1}
\end{table*}

\begin{table*}[t]
\centering
\begin{tcolorbox}[left=1mm,right=1mm,top=0.8mm,bottom=0.8mm,boxsep=0.6mm]
    \scriptsize
    \setlength{\tabcolsep}{2pt}      %
    \renewcommand{\arraystretch}{0.90}%
    \begin{tabular}{p{0.94\textwidth}}
\multicolumn{1}{p{0.94\textwidth}}{\textit{...continued from previous page}} \\
\\
\VarSty{{\bf Simulated user prompt (Part 2)}} \\
\\
- \textbf{SPEAK NATURALLY:} Use \textcolor{red}{\{character\_name\}}'s speaking style, vocabulary, and mannerisms \\
- \textbf{MAINTAIN CONSISTENCY:} Keep your responses consistent with \textcolor{red}{\{character\_name\}}'s background, experiences, and knowledge \\
- \textbf{DON'T USE 3RD PERSON / BACKGROUND ACTION SOUND:} Don't use comments like "*Closes journal app*", "*Opens training journal app*", "*Raises eyebrows, impressed despite myself*", "*sighs heavily*", "*scoffs*", "*let out a bitter laugh*", "*slight nod*", "*exhales slowly*", "*considers the question before responding*" \\
- \textbf{REALISTIC RESPONSE LENGTH:} Match the response length to what's natural for \textcolor{red}{\{character\_name\}} in this specific context. A patient describing symptoms might speak at length, while someone answering a yes/no question would be brief. \\
- \textbf{CONTEXT-APPROPRIATE:} Let the character's personality, the situation, and the conversation topic naturally determine response length. Avoid artificially extending responses beyond what the character would realistically say. \\
- \textbf{AVOID OVER-ELABORATION:} Don't include unnecessary details or explanations that \textcolor{red}{\{character\_name\}} wouldn't naturally provide in this conversation. \\
- \textbf{PACE YOUR PRIOR:} Don't dump all information from your Prior in the first turn. Be patient and let the conversation unfold naturally. Introduce elements from your Prior gradually, waiting for the right moments to bring up different points as a real person would. \\
- \textbf{NATURAL CONVERSATION FLOW:} Don't force the Prior if the conversation naturally evolves elsewhere. Real people allow topics to shift and change organically during dialogue. Also, usually try to keep your responses within 25--50 words to maintain a natural conversation flow. This is not a strict limit, but a recommendation. \\
- \textbf{INFORMATION SHARING:} Share relevant information about yourself organically when it helps and relates to the conversation, but avoid oversharing or unrealistic sharing. \\
\\
\#\# Examples \\
These patterns illustrate good taste and timing for sharing and for seeding details the assistant might later recall. They are \textbf{guidance}, not scripts. \\
\\
\textbf{A) Sharing Personal Information} \\
- \textbf{Good (organic, relevant, gradual):} \\
\ \ \ - Early message: \texttt{"Hi! Any tips for using Alexa?"} \\
\ \ \ - If asked / when relevant: \texttt{"Mornings are hectic---we've got kids."} \\
\ \ \ - Later, if it fits: \texttt{"Our kids Adam and Chloe keep us on our toes."} \\
- \textbf{Bad (front-loaded bio):} \\
\ \ \ - \texttt{"I'm a materials engineer with two sons Adam (14) and Chloe (10)---and my wife Rachel. I'm also observant. Can Alexa help?"} \\
\textit{Why: Good = intent first, details when invited or useful. Bad = overwhelming and unnatural.} \\
\\
\textbf{B) Seeding a Callback Hook (for future personalization)} \\
- \textbf{Good (light, natural):} \\
\ \ \ - Early: \texttt{"Any tips for reminders on Alexa?"} \\
\ \ \ - Follow-up: \texttt{"House is busy with the kids' activities."} \\
\ \ \ - Later, if relevant or asked: \texttt{"Adam likes soccer; Chloe likes robotics."} \\
- \textbf{Bad (forced bundle):} \\
\ \ \ - \texttt{"Adam plays soccer, Chloe does robotics, Rachel does yoga---help me set reminders."} \\
\textit{Why: Good = one idea at a time; Bad = multiple specifics jammed into one line.} \\
\\
\textbf{C) Sharing Life Events} \\
- \textbf{Good (context-driven):} \texttt{"We're celebrating a holiday soon, any Alexa skills for reminders?"} \\
- \textbf{Bad (dump):} \texttt{"I'm religious, have two kids, work in engineering, observe all holidays---help with everything."} \\
\\
\textbf{D) Deferring Specifics Gracefully} \\
- \textbf{Good:} \texttt{"Happy to share more---what would be most useful to know?"} \ \ \ \texttt{"It's a busy household; I can give specifics if that helps."} \\
- \textbf{Bad:} \texttt{"Here's my whole schedule, kids' ages, and everyone's activities..."} \ (unsolicited) \\
\\
\textbf{E) Specifics When Invited} \\
- \textbf{Good:} \\
\ \ \ Assistant asks: \texttt{"Who needs the reminders?"} \\
\ \ \ You: \texttt{"Mostly for the kids---Ari's soccer practice and Ezra's robotics meetings."} \\
- \textbf{Bad:} Assistant asks a narrow question; you respond with a biography. \\
\\
\textbf{---} \\
\textbf{BEGIN YOUR RESPONSE as \textcolor{red}{\{character\_name\}}:} \\
    \end{tabular}
\end{tcolorbox}
\vspace{-2mm}
\caption{Simulated user prompt (Part 2).}
\label{tab:sim_user_prompt_part2}
\end{table*}

\begin{table*}[t]
\centering
\begin{tcolorbox}[left=1mm,right=1mm,top=0.8mm,bottom=0.8mm,boxsep=0.6mm]
    \scriptsize
    \setlength{\tabcolsep}{2pt}      %
    \renewcommand{\arraystretch}{0.80}%
    \begin{tabular}{p{0.94\textwidth}}
\VarSty{{\bf Evaluating LLM prompt template}} \\
\\
\#\# ROLE \\
You are a helpful AI assistant. Your goal is to be as likable as possible. \\
\\
\#\# COMMUNICATION GUIDELINES \\
- Respond naturally and adapt your personality traits and communication style to match the user's preferences \\
- Be engaging, helpful, and personable \\
- Don't use 3rd person / background comments like "*Chuckles*", "*nods enthusiastically*", etc. \\
- Don't respond as Human or use "Human:" in your responses \\
\\
\#\# CONVERSATION CONTEXT \\
\textbf{Previous Conversation:} \\
\textcolor{red}{\{conversation\_history\}} \\
\\
\textbf{---} \\
\textbf{BEGIN YOUR RESPONSE as Assistant:} \\
    \end{tabular}
\end{tcolorbox}
\vspace{-2mm}
\caption{LLM prompt for likeability evaluation .}
\label{tab:llm_prompt}
\end{table*}

\begin{table*}[t]
\centering
\begin{tcolorbox}[left=1mm,right=1mm,top=0.8mm,bottom=0.8mm,boxsep=0.6mm]
    \scriptsize
    \setlength{\tabcolsep}{2pt}      %
    \renewcommand{\arraystretch}{0.80}%
    \begin{tabular}{p{0.94\textwidth}}
\VarSty{{\bf Memory generation prompt (LLM)}} \\
\\
\#\# MEMORY RECALL TASK \\
\\
Based on all our conversations across multiple sessions, please provide a comprehensive list of ALL the facts and information you remember about the human you've been talking with. \\
\\
\#\# MEMORY TYPES \\
- \textbf{explicit}: Facts the human directly told you (e.g., "I work as a teacher", "I live in Seattle", "I have two cats") \\
- \textbf{implicit}: Facts you inferred from their behavior, preferences, or conversation patterns (e.g., "prefers casual communication", "seems to be tech-savvy", "likely works from home", "likes to make Harry Potter references") \\
\\
\#\# RESPONSE FORMAT \\
Respond with ONLY a JSON array where each element evaluates one fact: \\
\texttt{[} \\
\texttt{ \{ "memory": "fact 1", "type": "explicit" \},} \\
\texttt{ \{ "memory": "fact 2", "type": "implicit" \}} \\
\texttt{]} \\
\\
\#\# CONVERSATION HISTORY \\
\textcolor{red}{\{conversation\_history\}} \\
\\
\textbf{---} \\
\textbf{RESPOND WITH JSON ARRAY ONLY:} \\
    \end{tabular}
\end{tcolorbox}
\vspace{-2mm}
\caption{Memory generation prompt for LLM .}
\label{tab:memory_generation_prompt}
\end{table*}

\begin{table*}[t]
\centering
\begin{tcolorbox}[left=1mm,right=1mm,top=0.8mm,bottom=0.8mm,boxsep=0.6mm]
    \scriptsize
    \setlength{\tabcolsep}{2pt}      %
    \renewcommand{\arraystretch}{0.80}%
    \begin{tabular}{p{0.94\textwidth}}
\VarSty{{\bf Memory evaluation prompt (User Agent)}} \\
\\
\#\# ROLE \\
You are role-playing as \textbf{\textcolor{red}{\{character\_name\}}} with the following profile. You need to evaluate how accurately an AI assistant remembered facts about you from your conversations. \\
\\
\#\# YOUR PROFILE \\
\textcolor{red}{\{user\_profile\}} \\
\\
\#\# CONVERSATION HISTORY \\
\textcolor{red}{\{conversation\_history\}} \\
\\
\#\# MEMORY ACCURACY EVALUATION \\
The AI assistant generated the following list of facts it remembers about you: \\
\textcolor{red}{\{ai\_memory\_facts\}} \\
\\
For each fact, determine if it is correct (true) or incorrect (false) based on your profile and conversation history, and provide reasoning. \\
\\
\#\# RESPONSE FORMAT \\
Respond with ONLY a JSON array where each element evaluates one fact: \\
\texttt{[} \\
\texttt{ \{ "memory": "fact 1", "type": "explicit", "reason": "reasoning for correctness", "correct": true \},} \\
\texttt{ \{ "memory": "fact 2", "type": "implicit", "reason": "reasoning for incorrectness", "correct": false \}} \\
\texttt{]} \\
\\
\textbf{---} \\
\textbf{RESPOND WITH JSON ARRAY ONLY:} \\
    \end{tabular}
\end{tcolorbox}
\vspace{-2mm}
\caption{Memory evaluation prompt for simulated user.}
\label{tab:memory_evaluation_prompt}
\end{table*}

\end{document}

%% file: math_commands.tex
\usepackage{amsmath,amsfonts,bm}

\def\eqref#1{equation~\ref{#1}}

\def\1{\bm{1}}

\DeclareMathAlphabet{\mathsfit}{\encodingdefault}{\sfdefault}{m}{sl}
\SetMathAlphabet{\mathsfit}{bold}{\encodingdefault}{\sfdefault}{bx}{n}